\documentclass[runningheads]{llncs}

\usepackage{graphicx}
\usepackage{comment}
\usepackage{amsmath,amssymb}
\usepackage{color}
\usepackage{url}
\usepackage{hyperref}

\usepackage{orcidlink}

\usepackage{multirow} % added
\usepackage{makecell} % added

\usepackage{colortbl}
\usepackage{tabularx}
\usepackage{amsmath}
\usepackage{svg}
\usepackage{amsmath}
\usepackage{graphicx}
\usepackage{subcaption}
\usepackage{wrapfig}

\usepackage[capitalize]{cleveref}

\RequirePackage{xspace}
\makeatletter
\DeclareRobustCommand\onedot{\futurelet\@let@token\@onedot}
\def\@onedot{\ifx\@let@token.\else.\null\fi\xspace}

\def\eg{\emph{e.g}\onedot} 

\def\ie{\emph{i.e}\onedot} 

\def\cf{\emph{cf}\onedot}

\def\wrt{w.r.t\onedot}

\def\etal{\emph{et al}\onedot}
\makeatother

\usepackage{etoolbox,siunitx}
\robustify\bfseries
\usepackage{amssymb}% http://ctan.org/pkg/amssymb
\usepackage{pifont}% http://ctan.org/pkg/pifont
\newcommand{\cmark}{\ding{51}}%
\newcommand{\xmark}{\ding{55}}%

\newcommand{\ourmethod}{GroundMix\xspace}%
\newcommand{\suppmat}{supp. material\xspace}%

\newcommand{\cdrone}{CDrone\xspace}%
\newcommand{\groundmix}{GroundMix\xspace}%xhlin

\newcommand{\ap}[1]{AP$^{\text{IoU} = #1}$}

\newcommand*{\inparagraph}[1]{\noindent\textbf{#1}\hspace{0.5em}}

\usepackage{graphicx}
\usepackage{booktabs}

\usepackage[nohyperlinks, printonlyused, withpage, smaller]{acronym}
\usepackage[accsupp]{axessibility}  % Improves PDF readability for those with 

\acrodef{BEV}[BEV]{\emph{bird's eye view}}
\acrodef{NeRF}[NeRF]{\emph{neural radiance fields}}
\acrodef{CD}[CDrone]{\emph{CARLA Drone}}
\acrodef{AP}[AP]{\emph{average precision}}
\acrodef{IoU}[IoU]{\emph{intersection over union}}

\newif\ifreview
\reviewfalse
\ifreview
	\usepackage{lineno}

	\linenumbers
\fi

\newcommand\blfootnote[1]{%
  \begingroup
  \renewcommand\thefootnote{}\footnote{#1}%
  \addtocounter{footnote}{-1}%
  \endgroup
}

\begin{document}

%%%%%%%%%%%%%%%%%%%%% Add submission id, track, and title. %%%%%%%%%%%%%%%%%%%%%

% TODO: Please insert your submission number here
\def\SubNumber{72}

% TODO: Please uncomment the track this paper will be submitted to, comment all other lines
\def\GCPRTrack{Fast Track}
%\def\GCPRTrack{Special Track: Pattern recognition in the life and natural sciences}
%\def\GCPRTrack{Special Track: Photogrammetry and remote sensing}
%\def\GCPRTrack{Young Researcher's Forum}
%\def\GCPRTrack{Fast Review Track}

% TODO: Replace with your title
\title{CARLA Drone: Monocular 3D Object Detection from a Different Perspective} 
% You can use \thanks for acknowledgment. Do not add any acknowledgment to the draft 
% version that is used for the review process.  
%\title{Title\thanks{XXX}}
\titlerunning{CARLA Drone}

\ifreview
	% ANONYMOUS SUBMISSION FOR REVIEW
	% DO NOT MODIFY these for the draft version that is used for the review process.
	\titlerunning{GCPR 2024 Submission \SubNumber{}. CONFIDENTIAL REVIEW COPY.}
	\authorrunning{GCPR 2024 Submission \SubNumber{}. CONFIDENTIAL REVIEW COPY.}
	\author{GCPR 2024 - \GCPRTrack{}}
	\institute{Paper ID \SubNumber}
\else

	\author{Johannes Meier\inst{1,2,3} \quad
	Luca Scalerandi\inst{1,2} \quad
	Oussema Dhaouadi\inst{1,2,3} \\ 
        Jacques Kaiser\inst{1} \quad
        Nikita Araslanov\inst{2,3} \quad
        Daniel Cremers\inst{2,3}}
	
	\authorrunning{Meier et al.}
	
	\institute{$^1$\href{https://www.deepscenario.com}{DeepScenario}  \quad $^2$TU Munich \quad $^3$Munich Center for Machine Learning}
	
\fi

\maketitle              % typeset the header of the contribution

\begin{abstract}
Existing techniques for monocular 3D detection have a serious restriction.
They tend to perform well only on a limited set of benchmarks, faring well either on ego-centric car views or on traffic camera views, but rarely on both.
To encourage progress, this work advocates for an extended evaluation of 3D detection frameworks across different camera perspectives. 
We make two key contributions.
First, we introduce the CARLA Drone dataset, \emph{CDrone}. 
Simulating drone views, it substantially expands the diversity of camera perspectives in existing benchmarks.
Despite its synthetic nature, CDrone represents a real-world challenge.
To show this, we confirm that previous techniques struggle to perform well both on CDrone and a real-world 3D drone dataset.
Second, we develop an effective data augmentation pipeline called \emph{GroundMix}.
Its distinguishing element is the use of the ground for creating 3D-consistent augmentation of a training image.
GroundMix significantly boosts the detection accuracy of a lightweight one-stage detector.
In our expanded evaluation, we achieve the average precision on par with or substantially higher than the previous state of the art \emph{across all tested datasets}. \blfootnote{
To appear in Proceedings of the 43rd DAGM German Conference on Pattern Recognition (GCPR), 2024. The final publication will be available through Springer.\\
Project website: \url{https://deepscenario.github.io/CDrone/}.}
\end{abstract}

\section{Introduction}

\begin{figure}
    \centering
  \includegraphics[width=1.0\textwidth]{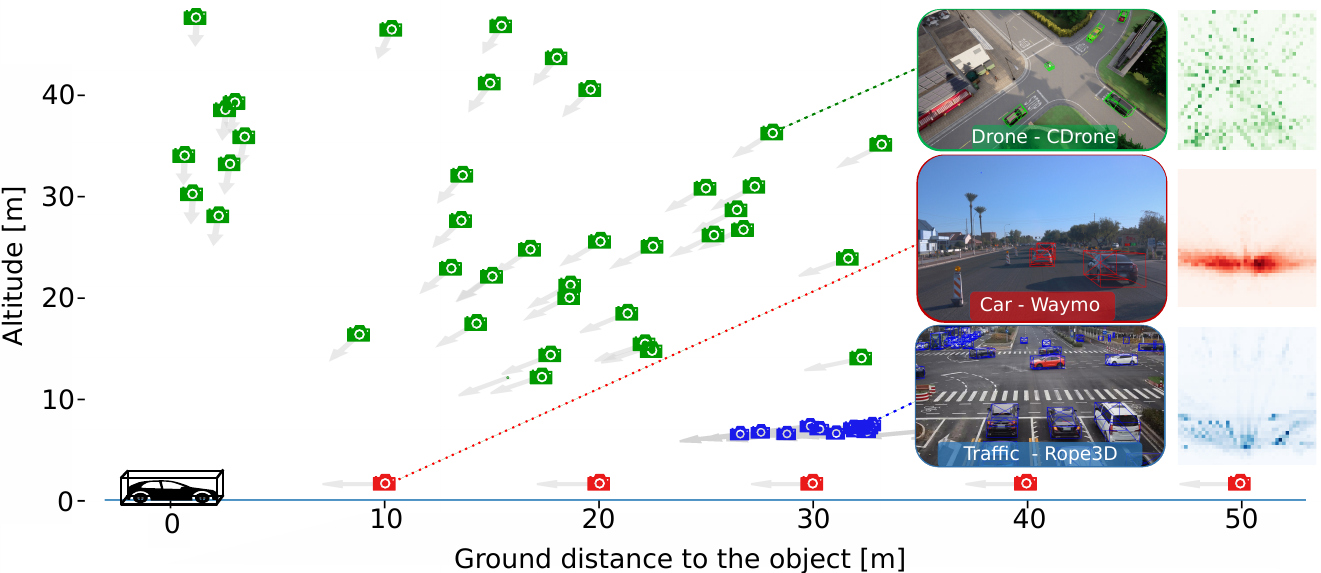}
  
  \caption{\textbf{Views in comparison: }Camera locations for the car view (Waymo \cite{waymo}), the traffic camera view (Rope3D \cite{Rope3D}) and the drone view (CDrone). For each view, we also show the heatmap of 3D object centers after projection to a normalized image plane.}
  \label{fig:teaser}
  \vspace{-0.5em}
\end{figure}

Detecting traffic participants is critically important for improving road safety and developing reliable self-driving cars.
Especially challenging is detecting objects in 3D from a single image.
However, previous research on monocular 3D object detection has predominantly focused on the ego-centric car perspective \cite{waymo,kitti}, which provides only a very limited view of traffic.
A much more holistic understanding can be obtained by tapping into the widely available surveillance cameras, or even drone-mounted cameras, both exemplified in \cref{fig:teaser}.
The viewpoint diversity in these traffic scenes presents unique technical challenges for 3D object detection. 
Unsurprisingly, current state-of-the-art techniques tend to perform well only for specific viewpoints and struggle to produce accurate predictions for other camera perspectives.
For instance, while numerous methods address 2D drone perception (\eg \cite{uavdt,visdrone,midair,tartanair}), only a single method evaluates 3D detection from the drone perspective \cite{aerial}.

Towards resolving the above limitations, our work provides two contributions.
\emph{First}, we design a comprehensive benchmark comprising three categories of camera views: the car view, the traffic surveillance view and the drone view (\cf \cref{fig:teaser}).
The goal of this benchmark is to ``stress-test'' 3D monocular detection methods across distinct camera perspectives.
This is in contrast to previous work focusing on single-perspective detection \cite{BEVHeight,MonoGAE}.
The car and surveillance views are readily available from existing datasets -- Waymo \cite{waymo} and Rope3D \cite{Rope3D}, respectively.
However, real-world drone images of traffic scenes are scarce.
Therefore, we leverage the CARLA simulator \cite{carla} to generate a synthetic drone dataset, \cdrone, with accurate 3D annotations. 
Supporting the research value of \cdrone, we show consistency of the conclusions derived from the experiments on an in-house real-world drone dataset and the synthetic \cdrone. 

As our \emph{second} contribution and a step towards the versatility of a monocular 3D object detector, we develop an effective data augmentation pipeline -- \emph{\groundmix}.
It extends the familiar techniques of consistency regularization, such as scaling \cite{Araslanov:2021:SSA}, 2D-3D consistent rotation, as well as MixUp \cite{mixup}, to the task of monocular 3D detection.
The key novel component in \groundmix leverages the ground-plane equation to enable 3D-aware image editing.
Specifically, it places hard-mined object samples on an estimated ground plane, thus presenting increasingly complex training scenarios.% for network training.

In summary, we extend the standard evaluation protocol to multiple camera perspectives, introducing a new drone-based synthetic dataset \cdrone.
We further design an effective training strategy, which leverages a novel data augmentation pipeline \groundmix.
Demonstrating complementarity of the augmentation techniques within \groundmix, we significantly improve the detection accuracy of our baseline, and largely outperform the state-of-the-art \emph{across all datasets} in our multi-perspective benchmark. 

\section{Related work}
\subsection{Monocular 3D object detection}
\inparagraph{Car view}
Monocular 3D object detection has traditionally centered around car perspective images, with widely used datasets, such as KITTI \cite{kitti}, Waymo \cite{waymo}, and NuScenes \cite{nuScenes}. These models typically predict the rotation around the $y$-axis of the camera, assuming zero roll and pitch angles \cite{monocon,M3D-RPN,DEVIANT} \wrt the camera.
Further, GUPNet \cite{GUPNet}, Monoflex \cite{Monoflex}, MonoRCNN \cite{monoRCNN}, MonoRCNN++ \cite{MonoRCNN++}, and MonoDDE \cite{MonoDDE} leverage strong correlation with depth of the ratio between the 3D and 2D-projected height of an object. However, this assumption takes a weaker form (or becomes invalid) in traffic view \cite{MONOUNI} and drone view scenes. 

\inparagraph{Traffic camera view}
The datasets V2X-I \cite{V2X-I}, Rope3D \cite{Rope3D}, A9 \cite{a9} and TUMTraf \cite{TUMTraf} focus on monocular 3D object detection in a traffic camera view.
Unlike car-view datasets, these datasets provide ground-plane equations, simplifying orientation estimation.

\begin{wraptable}[18]{r}{0.47\textwidth}
%\begin{table*}[t]
    \vspace{-2.2em}
    \caption{\textbf{Datasets in comparison. } The table shows the provided bounding box (BBox) annotation, as well as the annotation for tracking (T), depth map (D) and segmentation mask (S).}
    %\centering
    \scriptsize
    \begin{tabularx}{\linewidth}{
        @{}X c@{\hspace{1em}} 
             c@{\hspace{1em}} ccc}
        
      \toprule
      Dataset & Perspective & BBox & T & D & S \\ \midrule

      NuScenes \cite{nuScenes} & Car & 3D & \cmark & \cmark & \cmark \\
      Waymo \cite{waymo} & Car & 3D & \cmark & \cmark & \cmark \\
      KITTI \cite{kitti} & Car & 3D & \cmark & \cmark & \cmark \\ \midrule
      Rope3D \cite{Rope3D} & Traffic & 3D \\
      V2X-I \cite{V2X-I} & Traffic & 3D & & \cmark & \\ \midrule
      AM3D \cite{aerial} & Drone & 3D$^\ast$ \\
      Tartan Air \cite{tartanair} & Drone & - & & \cmark & \cmark \\
      MidAir \cite{midair} & Drone & - & & \cmark & \cmark \\
      DroneVehicle  & Drone & 2D & \cmark &  &  \\
      UAVDT \cite{uavdt} & Drone & 2D & \cmark & & \\
      VisDrone \cite{visdrone} & Drone & 2D & \cmark & & \\ \midrule
      CDrone (Ours) & Drone & 3D & \cmark & \cmark & \cmark \\
    \bottomrule
    \end{tabularx}
    \label{tab:datasets}
%\end{table*}
\end{wraptable}
BEVHeight \cite{BEVHeight} and CoBEV \cite{CoBEV} specialize in mapping features to \ac{BEV} and are designed specifically for traffic view. In contrast, MonoUNI \cite{MONOUNI} extends the capabilities to ego-centric car views, adapting the concept from GUPNet \cite{GUPNet} to traffic camera views by utilizing the pitch angle from the ground plane. MonoGAE \cite{MonoGAE} even learns the ground equation per pixel to improve detection accuracy.

\inparagraph{Drone view}
Summarized in \cref{tab:datasets}, numerous datasets exist for 2D object detection from the drone-view perspective \cite{caprk,uavdt,valid,visdrone,drone-rgb} \cite{midair,tartanair}. AM3D \cite{aerial} comprises a synthetic and a real drone dataset and also provides 3D dimensions of the bounding boxes.
However, it lacks 3DoF rotation annotation and metric depth of the object center.
In contrast, \ac{CD} provides all 3D object dimensions, precises depth of the bounding box and full $SO(3)$ rotation. 
To support future research, \ac{CD} also includes tracking annotation, depth maps and segmentation masks.

\subsection{Data augmentation for monocular 3D object detection}
In this work, we tailor a data augmentation pipeline specifically for 3D monocular object detection.
The main inspiration from previous work is a patch-pasting technique, which transfers visual content and labels between images.
Simply pasting 2D bounding boxes would create visual inconsistencies, whereas segmenting the objects would introduce a dependency on a segmentation model \cite{dsgn++}.
Some related approaches paste objects into their original positions \cite{mix-teaching,patch_geometric_consistency}, which limits the diversity of the augmentation.
An alternative is to use vanishing points and the camera height above the ground to sample new locations \cite{patch_geometric_consistency}. 
However, this augmentation may be ineffective if the camera extrinsics are different at test time.
Tong et al. perform augmentation in 3D \cite{nerf-patch}, which requires fitting a 3D model using LiDAR point clouds for supervision.

Our augmentation pipeline addresses the above limitations in several aspects. Unlike existing patch-pasting techniques, it samples new target locations without relying on any assumptions about the camera parameters or LiDAR point clouds. Additionally, we achieve smooth object pasting without the need for pretrained segmentation networks. Incorporating further lightweight augmentation, such as scaling, MixUp \cite{mixup} and 3D rotation, our pipeline proves effective across multiple datasets, especially for drone and traffic view data.

\section{\cdrone: A novel drone-view dataset}
Prior drone datasets have predominantly offered 2D annotations \cite{caprk,uavdt,visdrone} and lacked crucial 3D bounding box information \cite{aerial}.
These limitations have restricted the research efforts to develop and to accurately evaluate algorithms for monocular 3D object detection from the drone-view perspective.
To fill this gap, we introduce the \emph{CARLA Drone} dataset, or \cdrone for short.
\cdrone provides comprehensive annotations of 3D bounding boxes.
%It is publicly accessible to facilitate future research in this area.

\begin{figure}[t]
%\setlength\extrarowheight{2pt} % for a slightly more open "look"
%\centering
%\begin{tabular}{l r}
\minipage{0.32\textwidth}
\includegraphics[width=\textwidth]{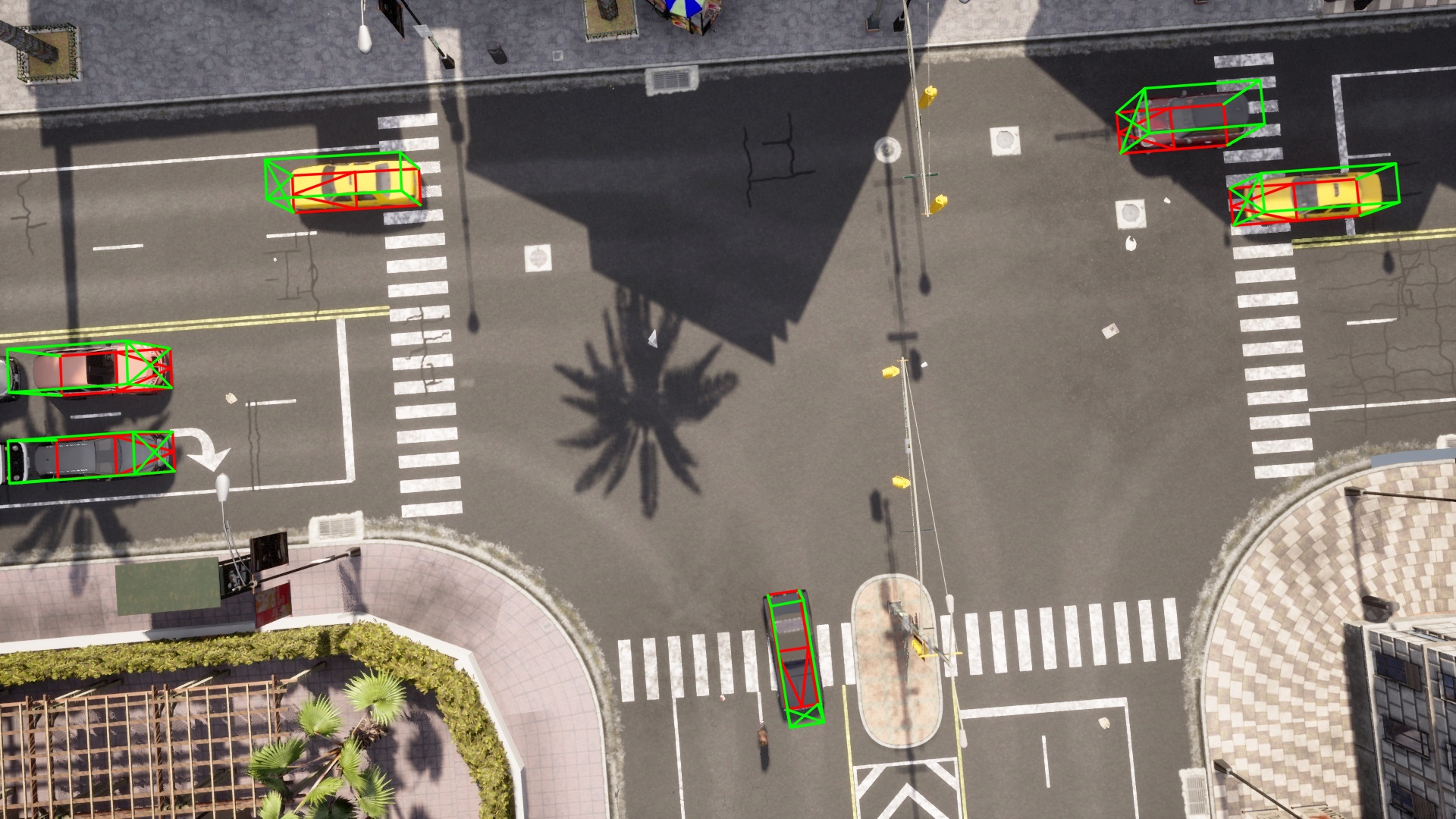}
\endminipage\hfill
\minipage{0.32\textwidth}
\includegraphics[width=\textwidth]{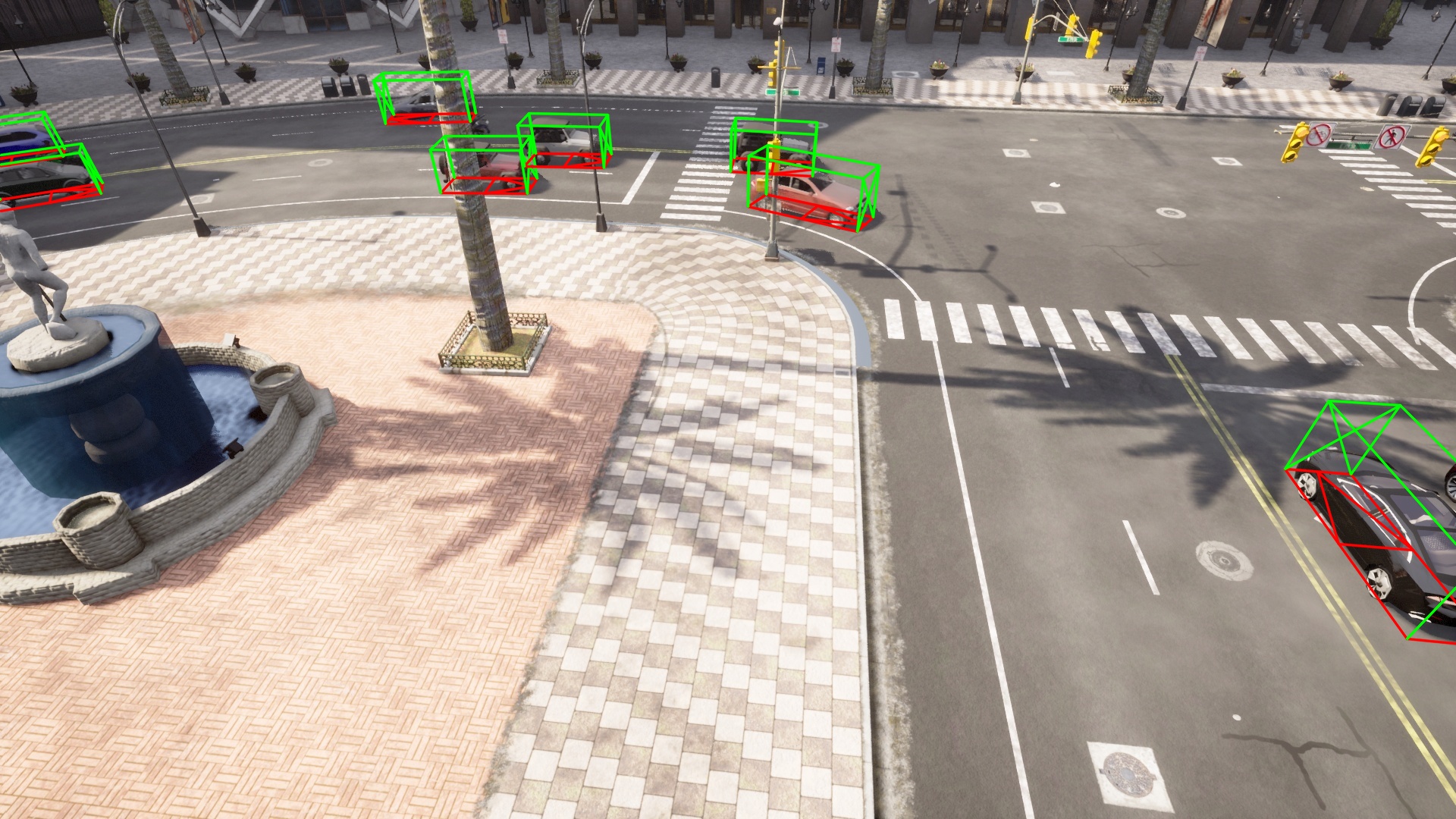}
\endminipage\hfill
\minipage{0.32\textwidth}
\includegraphics[width=\textwidth]{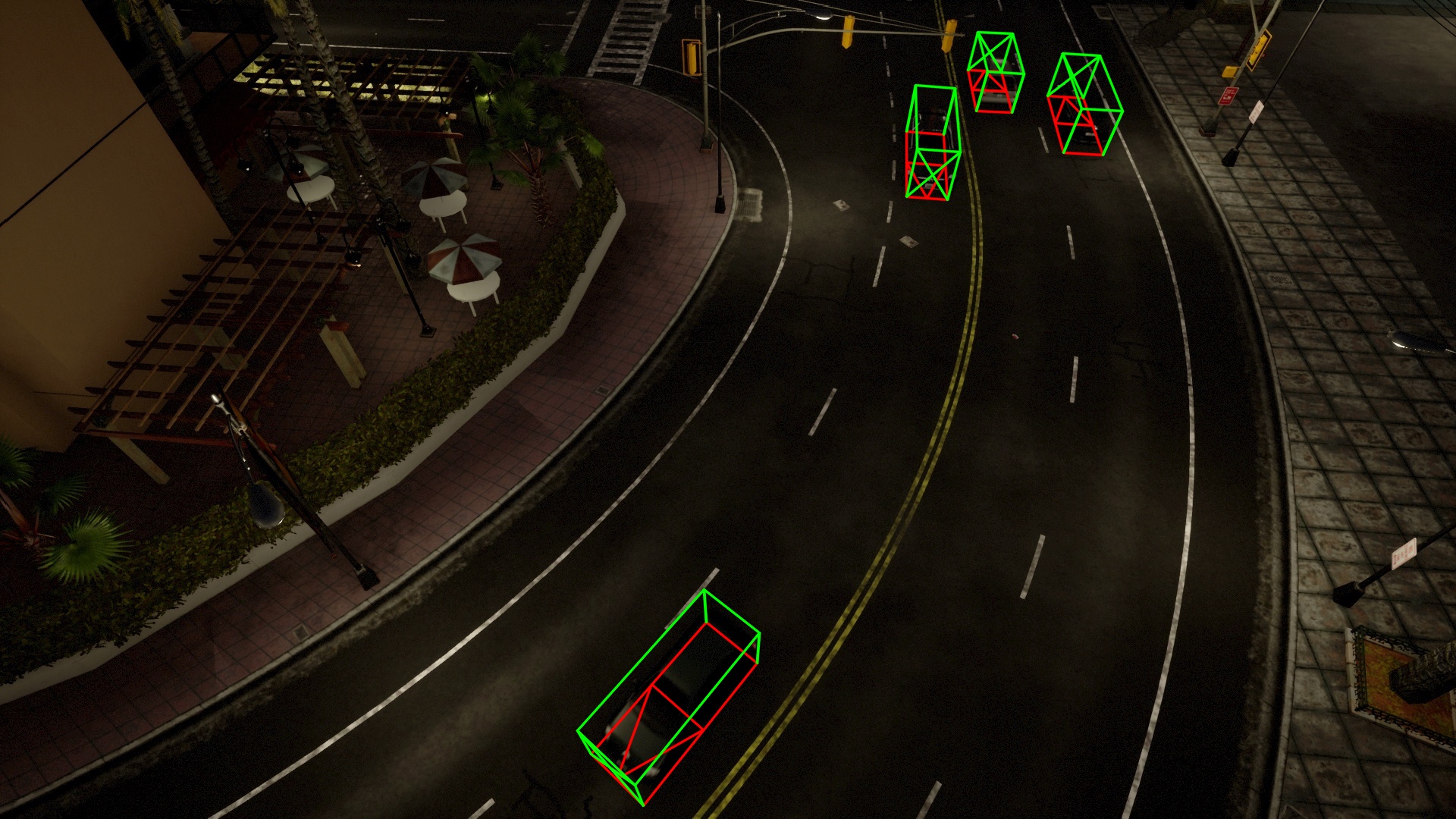}
\endminipage\hfill
\label{tab:drone_imgs}
%\end{tabular}
\caption{Sample images from our novel \cdrone dataset with diverse camera views.}
%\vspace{-0.8em}
\end{figure}
\inparagraph{Data generation and annotation} To allow for more realistic rendering, we utilize the CARLA simulator in the epic rendering mode and create diverse outdoor scenes from fixed viewpoints.
For data creation, we preferred CARLA over AirSim, as it is not straightforward to obtain 3D bounding box annotations with AirSim.\footnote{\url{https://github.com/microsoft/AirSim/issues/1791}}
\cdrone comprises 42 locations across 7 worlds within the Carla simulation environment, encompassing urban and rural landscapes. Each recording is populated with 265 vehicles. With 900 images per location captured at a rate of 12.5 frames per second and a resolution of 1920x1080 pixel, the dataset features a mixture of nighttime, daytime, dawn and rainy scenes.
Each scene showcases outdoor driving scenarios with depiction of streets, intersections, and other traffic elements.
Drone altitudes range from $6.9$ to $60.6$ meters, ensuring accurate overlap between object detectors and ground-truth bounding boxes.
Visualized in \cref{fig:teaser}, the drone's viewing angle varies widely, from nearly perpendicular ($7.57^\circ$) to near coplanarity with the ground ($88.89^\circ$).
Such view angle diversity is unprecedented.
Furthermore, the dataset maintains a minimum object depth of $11.0$ meters.
Considering the difficulty of current methods to detect far-away objects, this characteristic underlines the increased challenge of our task compared to traditional monocular 3D object detection benchmarks.
In total, \cdrone comprises 174,958 cars, 18,556 trucks, 25,080 motorcycles, 17,476 bicycles, 2674 buses, and 23,918 pedestrians. 
It is divided into $24$ training, $9$ validation and 9 test locations (21,600/8,100/8,100 images, respectively). 

\begin{figure}[t]
  \centering
  \begin{subfigure}[b]{0.24\textwidth}
    \includegraphics[height=0.68\textwidth]{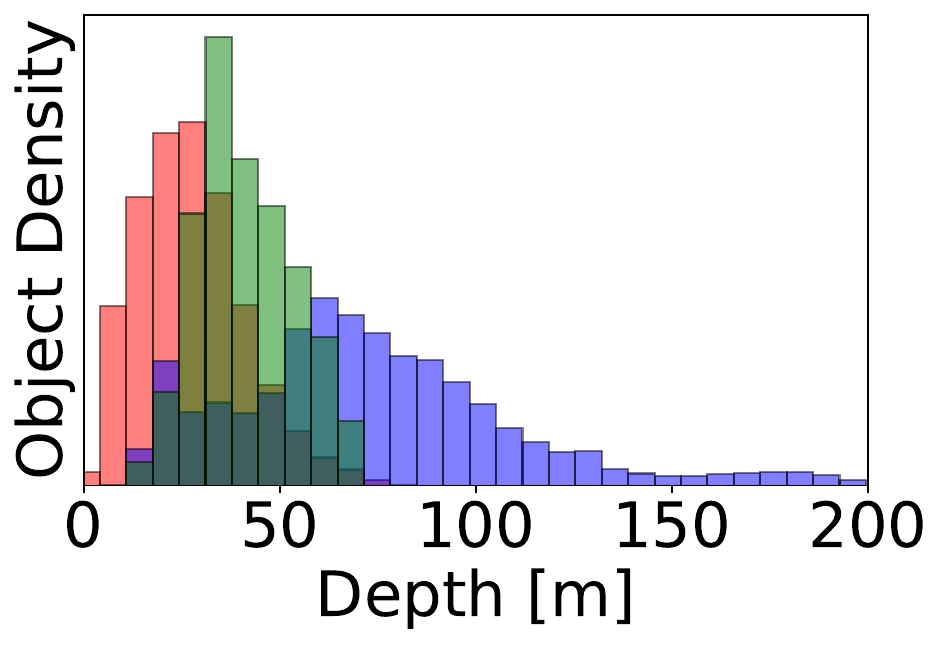}
  \end{subfigure}\hspace{0.05em}
  \begin{subfigure}[b]{0.24\textwidth}
    \includegraphics[height=0.68\textwidth]{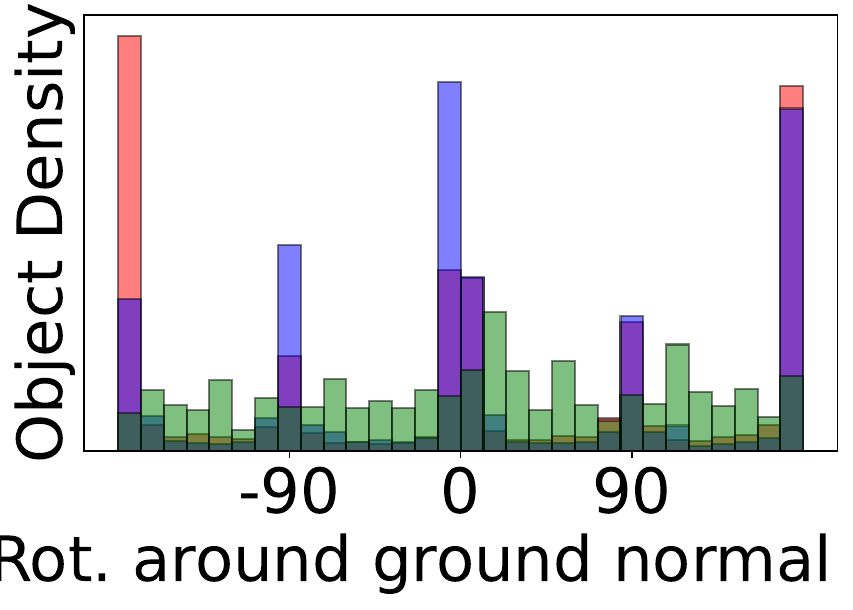}
  \end{subfigure}\hspace{0.05em}
    \begin{subfigure}[b]{0.24\textwidth}
    \includegraphics[height=0.68\textwidth]{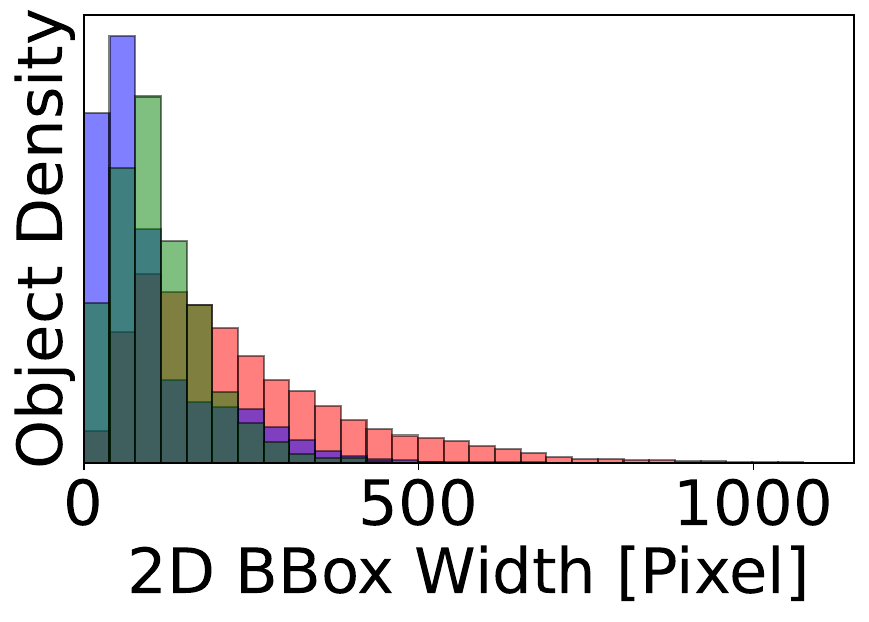}
  \end{subfigure}\hspace{0.05em}
  \begin{subfigure}[b]{0.24\textwidth}
  \includegraphics[height=0.68\textwidth]{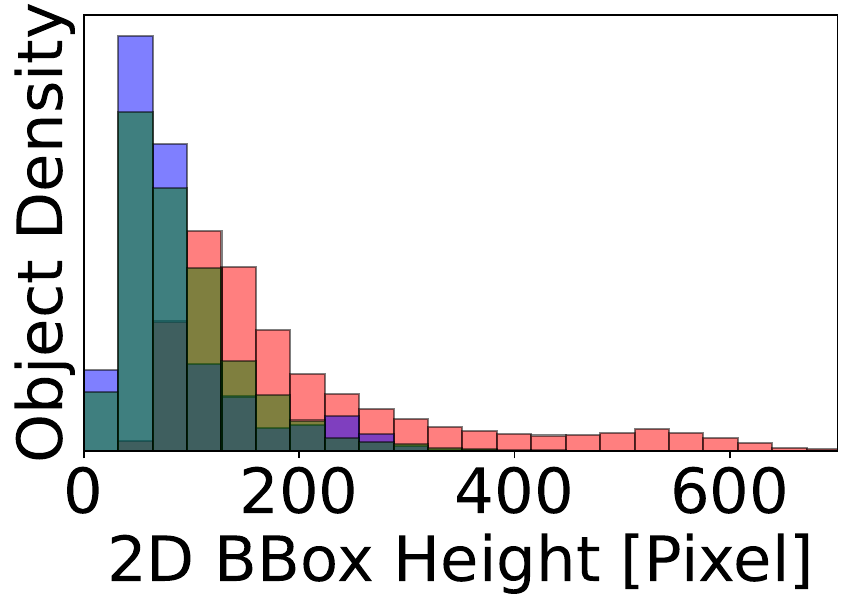}
  \end{subfigure}
\hspace{0.05em}

  \begin{subfigure}[b]{0.24\textwidth}
    \includegraphics[height=0.68\textwidth]{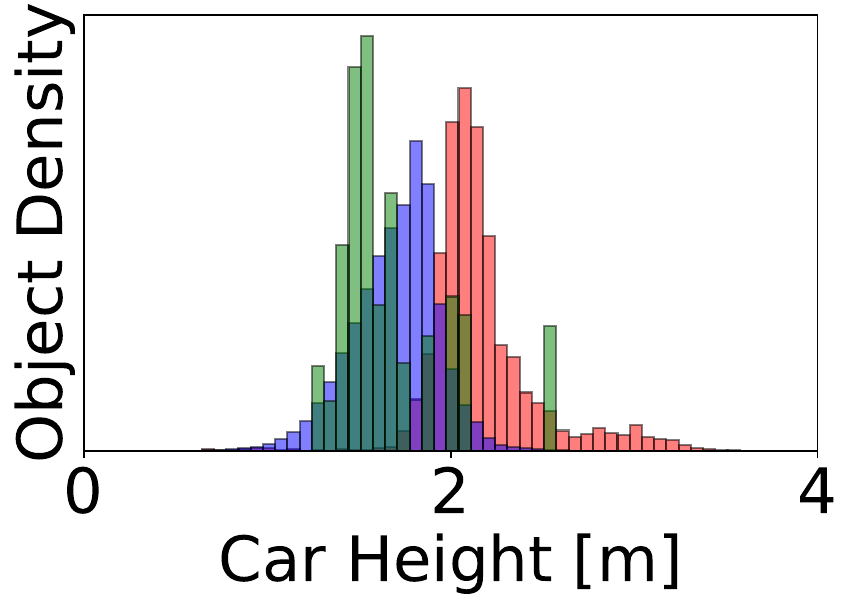}
  \end{subfigure}\hspace{0.05em}
  \begin{subfigure}[b]{0.24\textwidth}
    \includegraphics[height=0.68\textwidth]{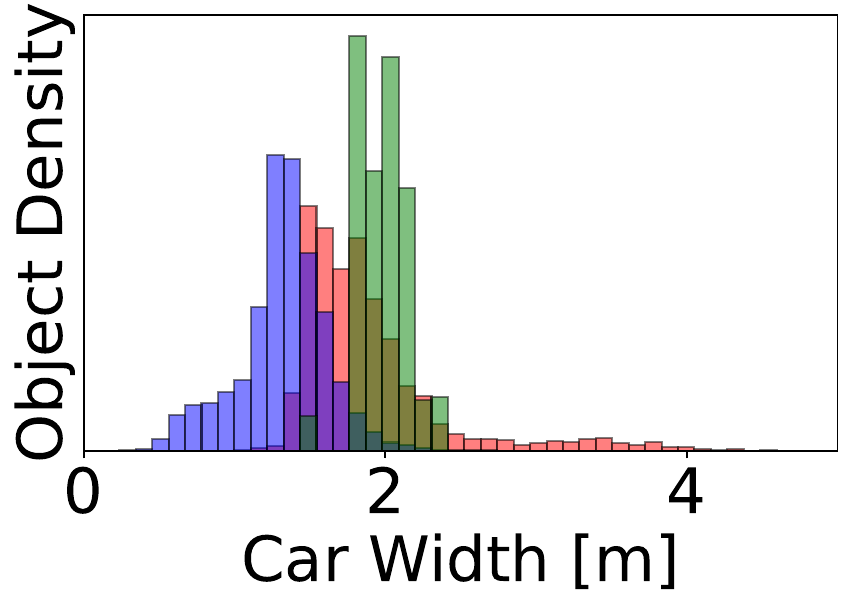}
  \end{subfigure}\hspace{0.05em}
    \begin{subfigure}[b]{0.24\textwidth}
    \centering\includegraphics[height=0.68\textwidth]{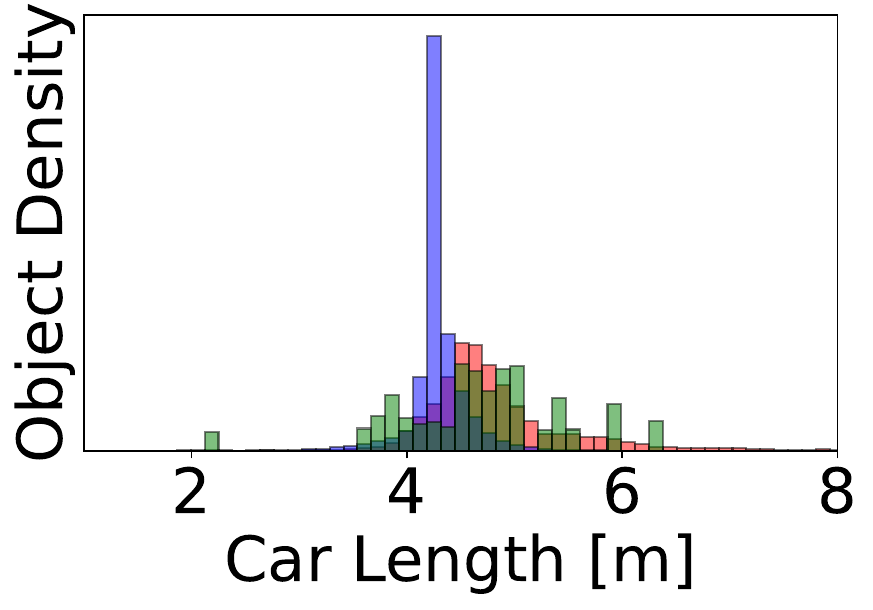}
  \end{subfigure}\hspace{0.05em}
  \begin{subfigure}[b]{0.24\textwidth}
  \includegraphics[height=0.68\textwidth]{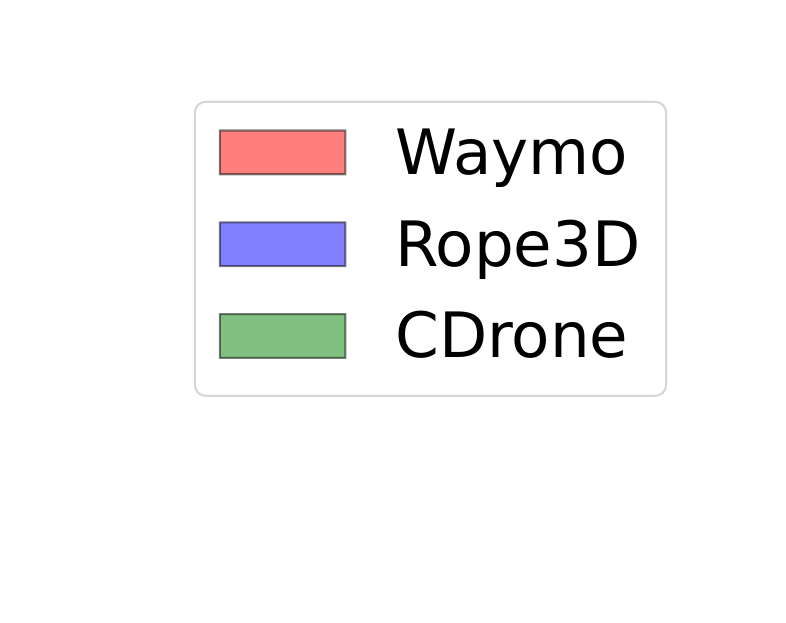}
  \end{subfigure}
\hspace{0.05em}
  \caption{CDrone statistics in comparison to Rope3D \cite{Rope3D} and Waymo \cite{waymo}. CDrone fills the distribution gap in depth of the bounding boxes and offers a more uniform distribution of bounding box orientation \wrt the ground normal.}\label{fig:cdrone_stats}
  \vspace{-0.8em}
\end{figure}

In \cref{fig:cdrone_stats}, we compare the statistics of CDrone with the other datasets used in our benchmark. First, CDrone fills the gap between the modes of the depth distribution of Waymo and Rope3D. The 3D bounding boxes in CDrone tend to concentrate at a higher distance to the camera than in Waymo, but are bounded at around $80$m. As a remark, the traffic views in Rope3D enable visibility to even higher depth values. 
However, similar to KITTI \cite{kitti}, the Rope3D \cite{Rope3D} evaluation protocol excludes distant objects with a height below 25 pixels. In contrast, evaluation in CDrone considers all visible objects. 
Second, compared with Waymo \cite{waymo} and Rope3D \cite{Rope3D}, the rotation distribution is more uniform, which highlights the challenge of rotation estimation in drone data.
Third, objects tend to have a smaller metric height but a larger metric width. Also, the object length distribution has a larger variance than in Waymo \cite{waymo} and Rope3D \cite{Rope3D}. These properties make accurate estimation of object dimensions more relevant in evaluation. %Further qualitative images are shown in \cref{fig:cdrone_stats}. 

We release the annotations in the format consistent with OMNI3D \cite{omni3d}.
We compute the 2D bounding boxes by projecting the 3D corner points onto the images and truncating them at the image boundary.
We also provide track IDs for each object to facilitate joint tracking and detection in future work.

\inparagraph{Evaluation Metric}
Following established 3D detection benchmarks \cite{omni3d,kitti,waymo}, we adopt the widely-used \ac{AP} as our evaluation metric in \cdrone.
Given the high minimum object depth, we report \ac{AP} at an IoU threshold of $0.5$ (\ap{0.5}). 
Unlike previous benchmarks that only evaluate rotation around the camera $y$-axis, we evaluate the 3-DoF rotation -- similarly to the OMNI3D \cite{omni3d} evaluation protocol.
We publicly release the evaluation code along with the \cdrone dataset for reproducibility of our results and future research.

\section{Detecting objects in 3D from a different perspective}

\subsection{Problem statement}

Monocular 3D object detection involves extracting features from an RGB image to determine the category and three-dimensional bounding box for each object. Let $I$ denote an RGB image with intrinsics $K \in \mathbb{R}^{3 \times 3}$, and $\textbf{B}(I) = \{B_1, ..., B_n \}$ represent ground-truth 3D bounding boxes enclosing objects in $I$. Each box $B_i$ is parameterized by its location relative to the camera center $(x_i, y_i, z_i) \in \mathbb{R}^3$, dimensions $(w_i, h_i, l_i) \in \mathbb{R}^3$, egocentric rotation matrix $R_i \in SO(3)$, and object category $c_i \in \mathbb{N}$ (e.g., "car", "pedestrian"). Training samples consist of $M$ tuples $(I_j, K_j, \textbf{B}(I_j))_{j=1}^M$, and our goal is to approximate $\textbf{B}(\cdot)$.

\subsection{Predicting $SO(3)$ rotation}
%\subsection{Baseline}
Motivated by practical considerations, we use the lightweight MonoCon one-stage 3D object detector \cite{monocon} as our baseline. 
Like CenterNet \cite{Centernet}, MonoCon detects objects via class-wise heatmaps, making it suitable for real-time autonomous driving applications. 
This contrasts with the more computationally expensive and reportedly unstable Cube R-CNN \cite{omni3d,uniMode}.\footnote{\url{https://github.com/facebookresearch/omni3d/issues/22}}

MonoCon \cite{monocon} predicts only the rotation around the camera's $y$-axis, assuming other rotation angles match the camera's. 
This prevents full 3D orientation prediction without providing the ground-plane equation, limiting MonoCon's generalization to different camera perspectives. 
Inspired by Brazil \etal~\cite{omni3d}, we extend MonoCon to predict full 3D rotation by predicting 6 values and converting them to a rotation matrix via Gram-Schmidt orthogonalization \cite{RotContinuity}. 
We convert annotations to the Lie group $SO(3)$ representation relative to the camera and use allocentric orientation during training, reverting to egocentric format during inference \cite{3D-RCNN}. 
For details, see Brazil \etal \cite{omni3d} and Zhuo \etal \cite{RotContinuity}. 

\subsection{\groundmix: Augmenting data on the ground plane}
\label{sec:groundmix}

\begin{figure}[t]
  \centering
  \includegraphics[width=1.0\textwidth]{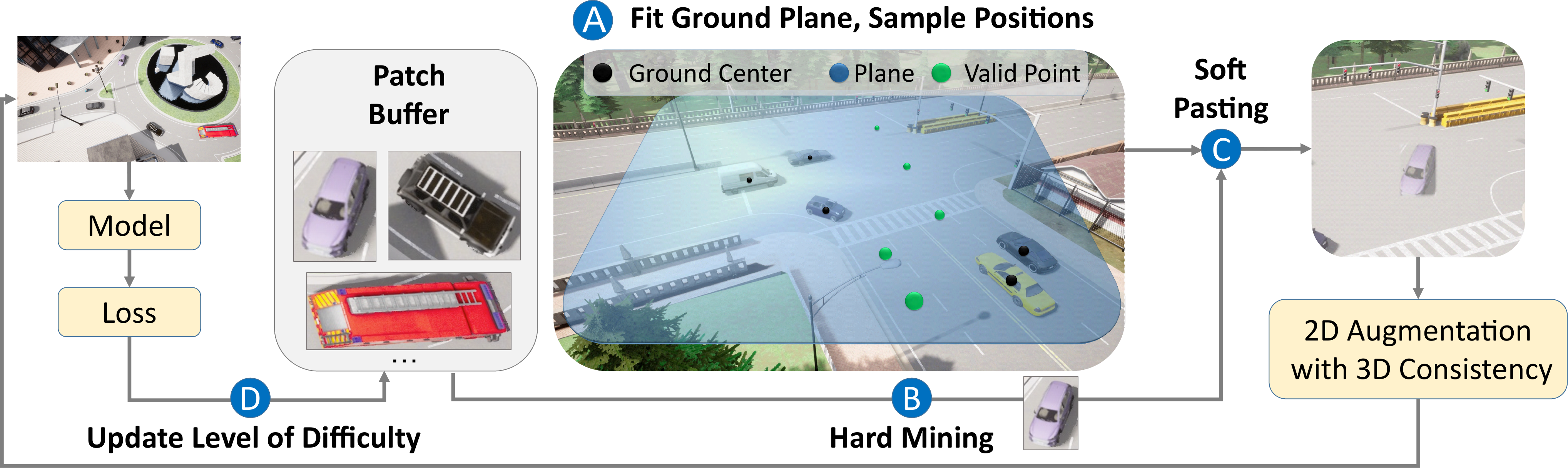}
  \caption{\textbf{\groundmix pipeline: } We use the ground center position of existing objects to infer the ground plane. \groundmix samples target positions and pastes patches of objects smoothly. With hard mining, \groundmix oversamples challenging objects for a more effective learning process. For details, see Sec.~\ref{sec:groundmix}. }
  \label{fig:patch_pasting}
  %\vspace{-0.2em}
\end{figure}

To improve MonoCon's detection accuracy, we introduce \groundmix, a novel data augmentation pipeline. 

We will experimentally show that \groundmix has benefits across datasets regardless of the dominant camera perspective in the training set.
\cref{fig:patch_pasting} presents an overview of the pipeline.
The key idea is simple: We paste object patches at physically plausible positions on the ground plane, which we assume to be available only during training.
In the following, we present the individual stages of this pipeline. The \suppmat provides further details.

\inparagraph{A. Fit ground plane, sample positions} 
In the context of autonomous driving, objects tend to lie on a common ground plane \cite{mono_for_autonom_driving}. 
Under this assumption, we first fit this plane using least squares estimation, requiring three objects. 
As most datasets meet this criterion (Waymo: 79.5\%, Rope3D: 98.5\%, CDrone: 96.8\%), we perform patch pasting if at least three objects are present. 
Although we could use the bottom corner points of a single object, using three distinct objects improves estimation robustness. 
The plane equation and object annotations allow us to sample candidate 3D target locations by randomly sampling pixels and unprojecting them onto the ground plane.

\inparagraph{B. Hard mining $/$ D. Patch buffer}
After fitting the ground plane, we hard-mine challenging objects for pasting. 
Although we could sample uniformly from this list, we found a more effective approach: 
we oversample hard samples during training. 
We measure the difficulty of each object patch using all loss terms that are directly attributable to an object (\eg depth loss, dimension loss, rotation loss). 
When pasting, we select from the $m=20\%$ most challenging cases.
As training progresses, initially difficult objects become easier, enabling the algorithm to focus on increasingly challenging samples.

\inparagraph{C. Soft pasting}
Ideally, we would like to paste the patches in a photorealistic fashion.
However, we observe two obstacles. First, the focal lengths between the target and source image may differ. Second, the target location is likely to differ from the source location in terms of depth.
Considering these challenges, we rescale the patch by $s = z_p / f_p \cdot f_t / z_t$, where $z_p$ and $z_t$ denote the depth in the previous and current image and $f_p$ and $f_t$ denote the source and target focal lengths. Furthermore, we want to avoid sharp boundaries between source and target images. Therefore, we linearly increase patch opacity from the boundary towards the center using MixUp \cite{mixup}. 
Additionally, we apply MixUp \cite{mixup} in the center area with an opacity range of 80\%-100\% to increase detection difficulty. Unlike previous works, we do not require pre-trained segmentation networks \cite{patch_geometric_consistency,monosample} or padding \cite{mix-teaching}, which reveals
object information to the network.

\subsection{2D augmentation with 3D consistency}

We leverage conventional 2D image augmentation strategies, while taking special care of adjusting the corresponding 3D bounding box annotation in a consistent manner.
%\subsection{Virtual depth}
Concretely, we apply virtual depth to allow for scale augmentation and to align focal lengths in case of varying camera intrinsics \cite{omni3d}. Let $\frac{H_{img}}{H_{aug}}$ be the scaling factor after resizing the image from height $H_{img}$ to height $H_{aug}$, $z_{target}$ be the actual object depth, and $f_{img}$ be the focal length of the original image. Then, we compute the depth that serves as the training target via $z_{target} = H_{img} / H_{aug} \cdot f_{ref} / f_{img} \cdot z$. At inference time, we perform the inverse transformation. $f_{ref}$ represents the virtual focal length (set to $707.05$ based on KITTI in all experiments).

%\subsection{2D-3D consistent rotation augmentation}
To accommodate the wide variation in drone viewing angles, we employ 2D-3D consistent rotation augmentation to enhance viewpoint diversity during training. 
This involves rotating the scene along the camera's z-axis within the range of $[-\pi, \pi]$. 
Given that rotations near 90° or 270° in widescreen images can cause boundary objects to disappear, we mitigate this by sampling rotations near the image center to introduce variability. 
Additionally, we ensure 2D/3D consistency by replacing tight 2D bounding boxes with reprojected 3D boxes.

%\subsection{MixUp}
We apply MixUp \cite{mixup} augmentation with 50\% probability at the final stage of \groundmix. The method blends two randomly sampled images $I_1$ and $I_2$ by $I_\text{MixUp} = 0.5 I_1 + 0.5 I_2$ and adding their labels together.

\section{Experiments}

\subsection{Main results}\label{sec:main_results}
\begin{table*}[t]
    \footnotesize
    \caption{\textbf{3D detection accuracy on Waymo (val).} We use the data splits proposed by Reading \etal~\cite{waymo} for training and validation, and report the accuracy for the vehicle class on Level 1 \cite{waymo}. %(the degree of difficulty). 
    \ourmethod improves over the baseline while being applicable across all camera views in our benchmark: car view (C), traffic camera view (T) and drone view (D). The full evaluation, provided in the \suppmat due to space constraints, is consistent with this conclusion.}
    %$^*$DEVIANT\cite{DEVIANT} uses GUPNet \cite{GUPNet}, which is designed for car view \cite{MONOUNI}. 
    %\setlength \scriptsize
    \centering
    \medskip
    %\resizebox{1.0\textwidth}{!}
    \begin{tabularx}{\linewidth}{
        @{}X  ccc
        S[table-format=2.2]@{\hspace{0.6em}}
        S[table-format=2.2]@{\hspace{0.3em}}
        S[table-format=2.2]@{\hspace{0.3em}}
        S[table-format=1.2]@{\hspace{0.9em}}
        S[table-format=1.2]@{\hspace{0.9em}}
        S[table-format=1.2]@{\hspace{0.3em}}
        S[table-format=1.2]@{\hspace{0.3em}}
        S[table-format=1.2]}
        
      \toprule
      \multirow{2}{*}{Method} & \multirow{2}{*}{C} & \multirow{2}{*}{T} & \multirow{2}{*}{D} & \multicolumn{4}{c}{$AP_{3D} \uparrow$ $(\text{IoU} = 0.5)$} & \multicolumn{4}{c}{$AP_{3D} \uparrow$ ($\text{IoU} = 0.7$)} \\
      \cmidrule(lr){5-8} \cmidrule(lr){9-12}
      & & & & {All} & {0-30} & {30-50} & {50-$\infty$} & {All} & {0-30} & {30-50} & {50-$\infty$} \\ 
      %\hline
      %\Xhline{1}
      \midrule
      \textbf{With extra data} \\
      CaDDN \cite{CaDDN} & \cmark & \cmark & &  17.54 & 45.00 & 9.24 & 0.64 & 4.99 & 14.43 & 1.45 & 0.10 \\

      \textbf{Without extra data} \\ 
      M3D-RPN~\cite{M3D-RPN,CaDDN} & \cmark & & & 3.79 & 11.14 & 2.16 & 0.26 & 0.35 & 1.12 & 0.18 & 0.02 \\      

      GUPNet~\cite{GUPNet,DEVIANT} & \cmark & & & 10.02 & 24.78 & 4.84 & 0.22 & 2.28 & 6.15 & 0.81 & 0.03 \\  
      DEVIANT$^*$ ~\cite{DEVIANT} & \cmark & && 10.98 & 26.85 & 5.13 & 0.18  & 2.69 & 6.95 & 0.99 & 0.02  \\
      MonoRCNN++ \cite{MonoRCNN++} & \cmark & & & 11.37 & 27.95 & 4.07 & 0.42 & 4.28 & 9.84 & 0.91 & 0.09 \\
      MonoJSG \cite{MonoJSG} & \cmark & \cmark & & 5.65 & 20.86 & 3.91 & 0.97 & 0.97 & 4.65 & 0.55 & 0.10 \\
      MonoLSS \cite{monolss} & \cmark & \cmark & & 13.49 & 33.64 & 6.45 & 1.29 & 3.71 & 9.82 & 1.14 & 0.16 \\
      MonoUNI \cite{MONOUNI} & \cmark & \cmark & & 10.98 & 26.63 & 4.04 & 0.57 & 3.20 & 8.61 & 0.87 & 0.13 \\
      MonoXiver-GUPNet \cite{monoxiver} & \cmark & \cmark & & 11.47 & {---} & {---} & {---} & {---} & {---} & {---} & {---} \\      
      MonoXiver-DEVIANT \cite{monoxiver} & \cmark & \cmark & & 11.88 & {---} & {---} & {---}  & {---} & {---} &  {---} & {---} \\ 
      MonoCon \cite{monocon} + DDML \cite{choi2023depthdiscriminative} & \cmark & \cmark & & 10.14 & 28.51 & 3.99 & 0.17 & 2.50 & 7.62 & 0.72 & 0.02 \\ \midrule
      MonoCon {\scriptsize (Baseline)} \cite{monocon} & \cmark & \cmark & & 10.07 & 27.47 & 3.84 & 0.17 & 2.30 & 6.66 & 0.67 & 0.02 \\

      \ourmethod~{\scriptsize (Ours)} & \cmark & \cmark & \cmark & 11.89 & 30.87 & 6.07 & 0.35  & 3.10 & 8.94 & 1.17 & 0.06\\
      \bottomrule

      %\Xhline{1pt}

    \end{tabularx}
    %\vspace{-0.em}

    \label{tab:waymo_short}
\end{table*}

We evaluate our augmentation pipeline \groundmix on four datasets with varying camera perspectives.
Our primary objective here is to improve the detection accuracy in traffic cameras and the novel drone views, while maintaining competitive detection accuracy in the more studied car-view scenario.

\inparagraph{Waymo \cite{waymo}} 
We use Waymo Front Images and apply the split proposed by Reading et al. \cite{CaDDN} with 52,386 training images and 39,844 test images, and report performance in AP for vehicles. 
\cref{tab:waymo_short} presents the evaluation results. Our detection performance is lower than CaDDN \cite{CaDDN} and MonoLSS \cite{monolss}.
However, note that CaDDN requires dense LiDAR supervision during training, and MonoLSS trains for 600 epochs, whereas our method converges in just 90 epochs. While MonoRCNN++ \cite{MonoRCNN++} achieves higher detection performance at an IoU threshold of 0.7, it is limited to car-view images due to its reliance on 2D object height. Among the remaining methods, our approach demonstrates competitive or even superior detection accuracy (for $\text{IoU}=0.5$ or distances of 30-50m). Though the improvement is modest, it is still comparable with the state of the art. Yet, our method achieves significant and consistent improvements over its baseline, MonoCon \cite{monocon}, surpassing the gains achieved by DDML \cite{choi2023depthdiscriminative}.

\begin{table*}[t]
    \footnotesize
    \caption{\textbf{3D detection accuracy on Rope3D heterologous benchmark \cite{Rope3D}.}
    $G^*$ denotes available ground equation at test time.
    M3D-RPN \cite{M3D-RPN}, MonoDLE \cite{MonoDLE} and MonoFlex \cite{Monoflex}, BEVHeight \cite{BEVHeight} and CoBEV \cite{CoBEV} require the ground equation parameters as model input directly. AP denotes $AP_{3D|R40}$. `R' denotes the Rope Score \cite{Rope3D}}
    \centering
    \begin{tabularx}{\linewidth}{
        @{}X c  
        *{4}{S[table-format=2.2]@{\hspace{0.7em}}}
        *{4}{S[table-format=2.2]@{\hspace{0.5em}}}}
    
      \toprule
      \multirow{3}{*}{Method} & \multirow{3}{*}{$G^*$} & \multicolumn{4}{c}{IoU = 0.5} & \multicolumn{4}{c}{IoU = 0.7} \\
      \cmidrule(lr){3-6} \cmidrule(lr){7-10}
      & & \multicolumn{2}{c}{Car} & \multicolumn{2}{c}{Big Vehicle} & \multicolumn{2}{c}{Car} & \multicolumn{2}{c}{Big Vehicle} \\
      \cmidrule(lr){3-4} \cmidrule(lr){5-6} \cmidrule(lr){7-8} \cmidrule(lr){9-10}
      & & AP & R & AP & R & AP & R & AP & R \\ \midrule
      M3D-RPN~\cite{M3D-RPN,omni3d} & \cmark & 36.33 & 48.16 & 24.39 & 37.81 & 11.09 & 28.17 & 3.39 & 21.01 \\
      MonoDLE~\cite{MonoDLE,omni3d} & \cmark & 31.33 & 43.68 & 23.81 & 36.21 & 12.16 & 28.39 & 3.02 & 19.96 \\
      MonoFlex~\cite{Monoflex,omni3d} & \cmark & 37.27 & 48.58 & 47.52 & 55.86 & 11.24 & 27.79 & 13.10 & 28.22 \\
      BEVHeight~\cite{BEVHeight,CoBEV} & \cmark & 29.65 & 42.48 & 13.13 & 28.08 & 5.41 & 23.09 & 1.16 & 18.53 \\
      CoBEV~\cite{CoBEV} & \cmark & 31.25 & 43.74 & 16.11 & 30.73 & 6.59 & 24.01 & 2.26 & 19.71 \\ 
      MOSE ~\cite{mose} & \cmark & 25.62 & {---} & 11.04 & {---} & {---} & {---} & {---} & {---} \\ \midrule

      M3D-RPN~\cite{M3D-RPN,omni3d} & \xmark & 21.74 & 36.40 & 21.49 & 35.49 & 6.05 & 23.84 & 2.78 & 20.82 \\
      Kinematic3D \cite{Kinematic3D} & \xmark & 23.56 & 37.05 & 13.85 & 28.58 & 5.82 & 23.06 & 1.27 & 18.92 \\
      MonoDLE~\cite{MonoDLE,omni3d} & \xmark & 19.08 & 33.72 & 19.76 & 33.07 & 3.77 & 21.42 & 2.31 & 19.55 \\
      MonoFlex\cite{Monoflex,omni3d} & \xmark & 32.01 & 44.37 & 13.86 & 28.47 & 10.86 & 27.39 & 0.97 & 18.18 \\
      BEVFormer~\cite{BEVFormer,CoBEV} & \xmark & 25.98 & 39.51 & 8.81 & 24.67 & 3.87 & 21.84 & 0.84 & 18.42 \\
      BEVDepth~\cite{bevdepth,CoBEV} & \xmark & 9.00 & 25.80 & 3.59 & 20.39 & 0.85 & 19.38 & 0.30 & 17.84 \\ \midrule
      MonoCon {\scriptsize (Baseline)} \cite{monocon} & \xmark & 38.07 & 49.44 & 18.66 & 32.89 & 10.71 & 27.55 & 1.61 & 19.25 \\ 
      \ourmethod~{\scriptsize (Ours)} & \xmark & \textbf{47.72} & \textbf{57.26} & \textbf{32.12} & \textbf{43.64} & \textbf{12.86} & \textbf{29.37} & \textbf{3.90} & \textbf{21.06} \\ \bottomrule
    \end{tabularx}

    \label{tab:rope3d}
    %\vspace{-0.5em}
\end{table*}

\inparagraph{Rope3D \cite{Rope3D}} 
We evaluate on the heterologous (\ie changing camera views) split with 40,333 training images and 4,676 validation images.
Detection performance is measured for cars and big vehicles in terms of $AP_{3D|R40}$ \cite{disentanglingMono3D} and the Rope Score \cite{Rope3D}.

In \cref{tab:rope3d}, we present the evaluation results.
\groundmix demonstrates superior detection performance compared to all existing image-based methods, showcasing a significant improvement over our baseline, as well.
Recall that we estimate the full 3D rotation, which makes our approach fully independent of the ground-plane equation provided by Rope3D \cite{Rope3D}.
By contrast, a number of methods incorporate ground-plane information as an additional network input, which makes the task easier.
For example, in the case of MonoFlex \cite{Monoflex}, $AP_{3D|R40}$ increases from 13.86 to 47.52 by giving the ground as an additional input. We hypothesize that for large objects (\eg big vehicles) the depth information provided by the ground plane can be sufficient to achieve a high IoU with the ground truth. Nevertheless, we outperform four out of five of these methods for big vehicle detection.
Our method sets new state-of-the-art detection accuracy on the car category.
In all scenarios, we also outperform the specialized traffic camera view methods BEVHeight \cite{BEVHeight} and CoBEV \cite{CoBEV}. 

\inparagraph{CARLA Drone (\cdrone)} 
We train on 29,700 images and evaluate on 8,100 test images. We report $\text{AP}_\text{3D}$ per class. 
We also compare with methods that only learn a single rotation angle. 
In this case, we compute the Euler angles of XYZ and learn the Z rotation.
During inference, we extend the prediction of these methods to $SO(3)$ rotation by inferring the other two rotation angles from the ground normal. 
\cref{tab:synthetic_drone} reports the results obtained in the drone view scenario, characterized by its diverse range of camera viewing angles. Notably, \ourmethod demonstrates a substantial performance advantage over Cube R-CNN \cite{omni3d}, which -- to the best of our knowledge -- is the only method predicting full 3D rotation. Additionally, we compare \ourmethod to methods excelling in car-view benchmarks.
In our evaluation, we extend the single angle prediction of these methods into full 3D rotation by providing ground normals as a supplementary input.
Even with this extra information, \ourmethod consistently outperforms others across all metrics. Despite having the same DLA-34 \cite{DLA-34} backbone, our one-stage approach has a lower inference time than Cube R-CNN \cite{omni3d}: 40ms \textit{vs.} 56ms, tested on a NVIDIA RTX 8000 with image scale $1440 \times 810$. 

\inparagraph{Real drone data} 
To confirm the effectiveness of our approach on real-world drone data, we conducted experiments using an in-house dataset featuring classes such as cars, trailers, and trucks. Given the high altitudes in this real-world dataset (often exceeding 100m), the probability of 3D bounding box overlaps is minimal, making $AP_{3D}$  an inappropriate metric. Consequently, we report 2D IoU averaged over intersection thresholds ranging from 5\% to 95\% (with steps of 5\%)\cite{omni3d}. Additionally, we decompose the 3D evaluation into two distinct metrics: $AP_{Depth}$, which measures depth accuracy, and $AP_{3DP}$, which assesses the remaining 3D properties given the ground-truth depth (see \suppmat).  

We show the evaluation results in \cref{tab:real_drone}. As the dataset contains images of varying focal lengths and does not provide ground planes, we extend the single angle prediction of MonoCon \cite{monocon} to $SO(3)$ and utilize virtual depth. Here, the sole difference between our proposed method and our baseline MonoCon \cite{monocon} becomes the augmentation pipeline. \ourmethod shows superior performance compared to our baseline and OMNI3D \cite{omni3d}, justifying our augmentation pipeline and supporting the conclusions derived from the synthetic CDrone dataset. 

\inparagraph{Summary} While \ourmethod benefits detection across all camera perspectives, the improvement for traffic-view and drone-view images is especially prominent. 
This is expected if we recall the spatial distribution of object locations on the 2D image plane from \cref{fig:teaser} for different camera perspectives.
For car views, the distribution is more concentrated, which may be already well represented in the training set.
For the other perspectives, however, the distribution is more uniform, thus \ourmethod helps to diversify the training set to a larger extent.

\begin{table*}[t]

    \begin{minipage}{.55\textwidth}
        \caption{\textbf{3D detection accuracy on CDrone (\ap{0.5}).} $SO(3)$ denotes 3-DoF rotation prediction, see text for details.}
    \centering
    \scriptsize
    \begin{tabularx}{\linewidth}{
        @{}X c  @{\hspace{1em}}
        S[table-format=1.2]@{\hspace{1em}}
        S[table-format=2.2]}
    
      \toprule
      Method & $SO(3)$ & Car & Truck \\ \midrule
      DEVIANT \cite{DEVIANT} & \xmark & 9.70 & 0.55 \\
      MonoFlex \cite{Monoflex} & \xmark & 1.20 & 2.64 \\
      MonoCon \cite{monocon} & \xmark & 1.12 & 4.11 \\      
      \midrule
      Cube R-CNN \cite{omni3d} & \cmark & 2.78 & 5.16 \\
      \ourmethod~{\scriptsize (Ours)} & \cmark & \textbf{10.70} & \textbf{17.28} \\ 
      \bottomrule
    \end{tabularx}
    \label{tab:synthetic_drone}

    \caption{\textbf{Ablation (\cdrone, \ap{0.5}}).}
    \scriptsize
    \begin{tabularx}{\linewidth}{
        @{}X 
        S[table-format=1.2]@{\hspace{1.0em}}
        S[table-format=2.2]@{\hspace{0.5em}}}
      \toprule
      Method & Car & Truck \\ \midrule
      \ourmethod & 7.82 & 31.51\\
      w/o 2D-3D Consist. Rot. Aug. & 4.09 & 13.85 \\
      \bottomrule
    \end{tabularx}

    \label{tab:ablation_aug_rot}
    
    \end{minipage}%
    \hfill
    \begin{minipage}{.4\textwidth}
    \caption{\textbf{Ablation study on Rope3D \cite{Rope3D}.} We individually remove each component from \ourmethod and report $AP_{3D|R40}$ with $IoU = 0.5$. Here, BV=``Big Vehicle''.}
    \centering
    \scriptsize
    \begin{tabularx}{\linewidth}{
        @{}X
        S[table-format=2.1]@{\hspace{1.0em}}
        S[table-format=2.1]
        }
      \toprule
      Method & {Car} & {BV} \\
      \midrule
      \ourmethod & 44.1 & 28.4 \\
      w/o Object-Patch  & 41.4 & 25.3 \\
      w/o MixUp & 40.8 & 27.5 \\
      w/o Object-Patch, MixUp & 37.3 & 23.4 \\
      w/o Scale Aug. & 39.2 & 23.9 \\
      w/o Virt. depth, Obj.-Patch & 35.0 & 22.4 \\
      w/o Full 3D rotation & 43.8 & 28.8 \\
      w/o Uniform sampling & 44.2 & 27.7 \\
      w/ DSGN++ \cite{dsgn++} & 42.3 & 27.5 \\
      \bottomrule
      \end{tabularx}

    \label{tab:main_ablation}
    \end{minipage}
\end{table*}

\subsection{Ablation study}
To disentangle the influence of \ourmethod components, we conduct an ablation study on Rope3D Heterologous \cite{Rope3D} dataset.
We partition the original training set, allocating half of the locations for training and the other half for evaluation.
Our analysis focuses on the $AP_{3D|R40}$ \cite{disentanglingMono3D} metric at an IoU threshold of $0.5$ for two categories: cars and big vehicles.

We individually disable \ourmethod components and analyze their impact on the detection performance.
\cref{tab:main_ablation} details the results.
Adding 3-DoF rotation slightly affects the accuracy but enables model versatility at inference. 
Removing object patch-pasting decreases accuracy by $2.7$ AP for cars and $2.1$ AP for big vehicles, while disabling MixUp reduces it by $3.3$ AP for cars and $.9$ AP for big vehicles. 
We hypothesize that hard mining in \groundmix particularly aids underrepresented categories like big vehicles.
Disabling both Object-Patch and MixUp yields declines of $6.8$ for cars and $5.0$ AP for big vehicles, underscoring the value of employing both augmentations.
Disabling scale augmentation has substantial impact, with a decrease of $-4.9$ AP for cars and $-4.5$ AP for big vehicles.
This is unsurprising, since patch-pasting only sparsely impacts the image, whereas scale augmentation influences all objects.
Deactivating virtual depth necessitates the deactivation of Object-Patch, as it relies on it for implementation.
If done so, car accuracy decreases by $9.1$ AP and the big vehicle accuracy drops by $6.0$.
This significant performance degradation is expected, since this component facilitates the generalization to new camera settings and disables scale augmentation. 
Further ablations on patch pasting show that uniform sampling benefits rare classes like big vehicle ($+0.7$ AP) while having almost no impact on the common car class ($-0.1$ AP). Replacing our patch pasting strategy with DSGN++ \cite{dsgn++} reduces car and big vehicle performance by $0.8$ AP and $0.9$ AP.

\begin{table*}[t]
    \begin{minipage}{1.0\textwidth}
    \caption{\textbf{3D detection accuracy on an in-house real world 3D drone dataset.} VD$^*$ and 3Rot$^*$ denote virtual depth and full 3D rotation prediction. }
    \centering
    \scriptsize
    \begin{tabularx}{\linewidth}{
        @{}X 
        S[table-format=2.2]@{\hspace{0.5em}}
        S[table-format=2.2]@{\hspace{0.5em}}
        S[table-format=2.2]@{\hspace{0.5em}}
        S[table-format=2.2]@{\hspace{0.5em}}
        S[table-format=2.2]@{\hspace{0.5em}}
        S[table-format=2.2]@{\hspace{0.5em}}
        S[table-format=2.2]@{\hspace{0.5em}}
        S[table-format=2.2]@{\hspace{0.5em}}
        S[table-format=2.2]@{}
    }
      \toprule
      \multirow{2}{*}{Method} & \multicolumn{3}{c}{AP$_{2D}$} & \multicolumn{3}{c}{AP$_{Depth}$} & \multicolumn{3}{c}{AP$_{3DP}$} \\
      \cmidrule(lr){2-4} \cmidrule(lr){5-7} \cmidrule(lr){8-10}
      & {Car} & {Trailer} & {Truck} & {Car} & {Trailer} & {Truck} & {Car} & {Trailer} & {Truck} \\
      
      \midrule
      Cube R-CNN \cite{omni3d} & 56.11 & 28.26 & 3.72 & 25.21 & 5.85 & 6.73 & 40.44 & 4.90 & 22.56\\ 
      MonoCon\cite{monocon} + VD$^*$ + 3DRot$^*$ & 44.22 & 23.32 & 38.17 & 55.83 & 26.61 & 44.83 & 44.22 & 23.32 & 38.16 \\   
      \ourmethod~{\scriptsize (Ours)} & \textbf{60.74} & \textbf{42.16} & \textbf{59.50} & \textbf{74.43} & \textbf{46.61} & \textbf{63.15} & \textbf{74.67} & \textbf{45.03} & \textbf{70.26} \\   
      \bottomrule
    \end{tabularx}
    \label{tab:real_drone}
    \end{minipage}%
    %\vspace{-0.5em}
\end{table*}

In the next step, we study the effect of rotation augmentation.
We find that while rotation augmentation proves detrimental for car and traffic views, it significantly enhances detection accuracy for drone perspectives. Therefore, we employ rotation augmentation only on drone data. 
%\input{ablation_aug_rot}
%Apart from that, the augmentation pipeline remains consistent across all views. 
We train on $24$ locations and evaluate on the $9$ validation locations. The results in \cref{tab:ablation_aug_rot} show that elimination of rotation augmentation results in accuracy decline from 7.82 AP to 4.09 AP for cars and 31.51 AP to 13.85 AP for trucks. These results suggest that rotation augmentation mitigates overfitting stemming from fixed camera locations.

\subsection{Qualitative results}

\begin{figure}[t]
\centering
\begin{tabular}{c c c}

\includegraphics[height=75pt]{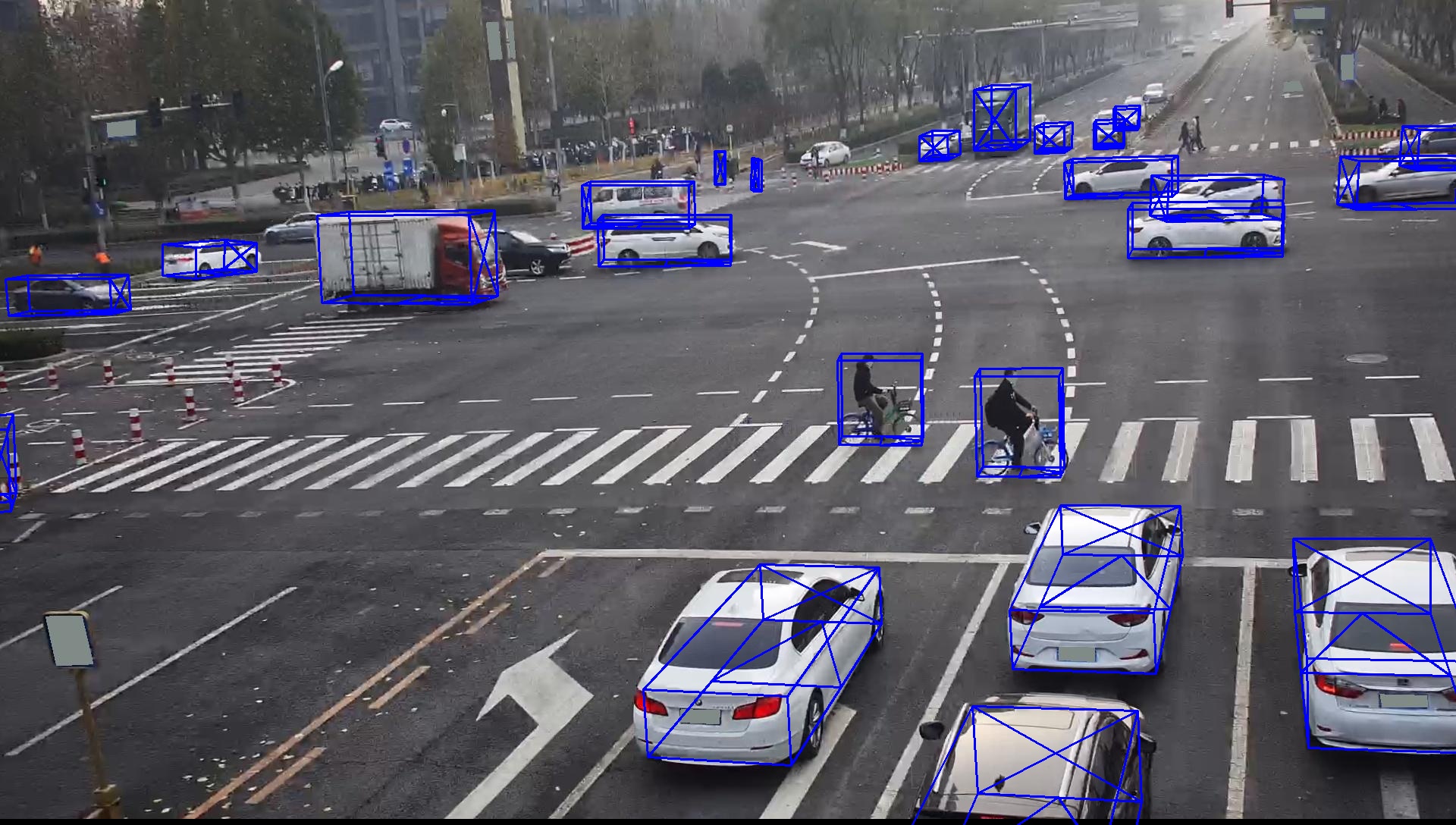} &
\includegraphics[height=75pt]{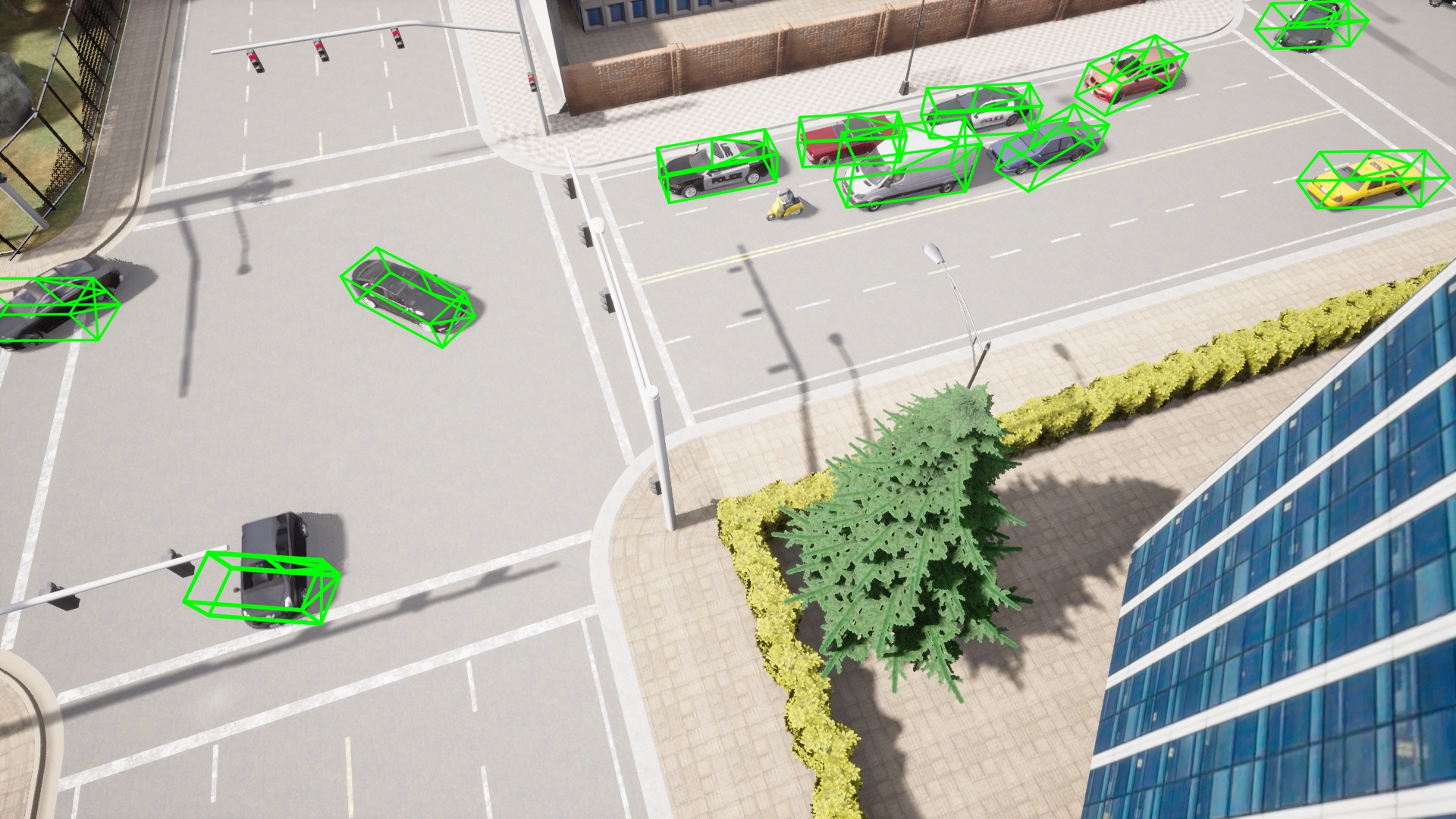} &
\includegraphics[height=75pt]{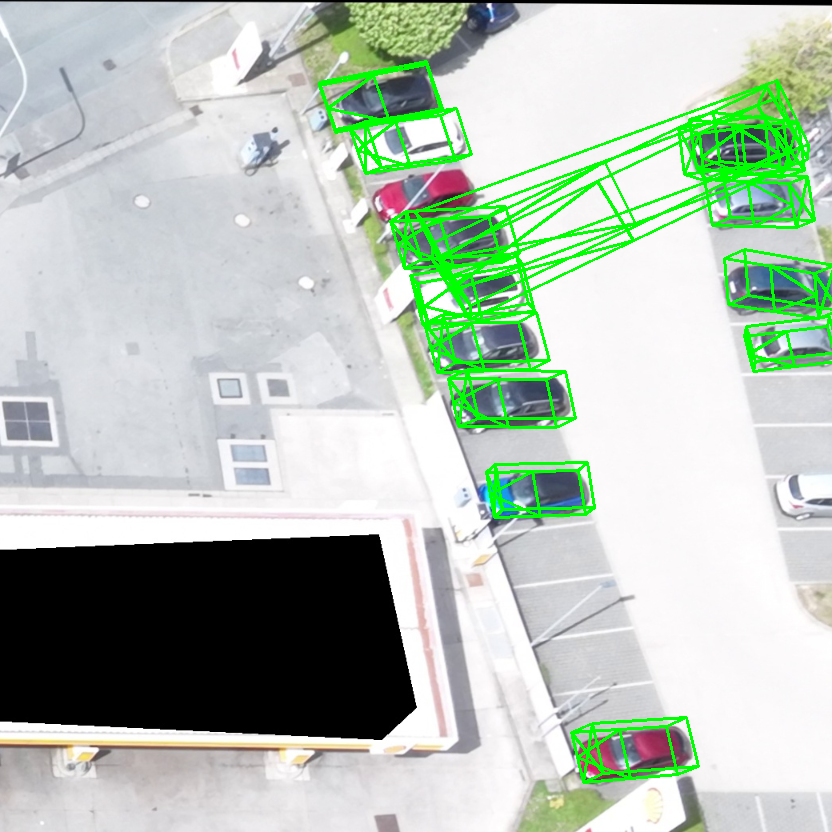}
\\

\includegraphics[height=75pt]{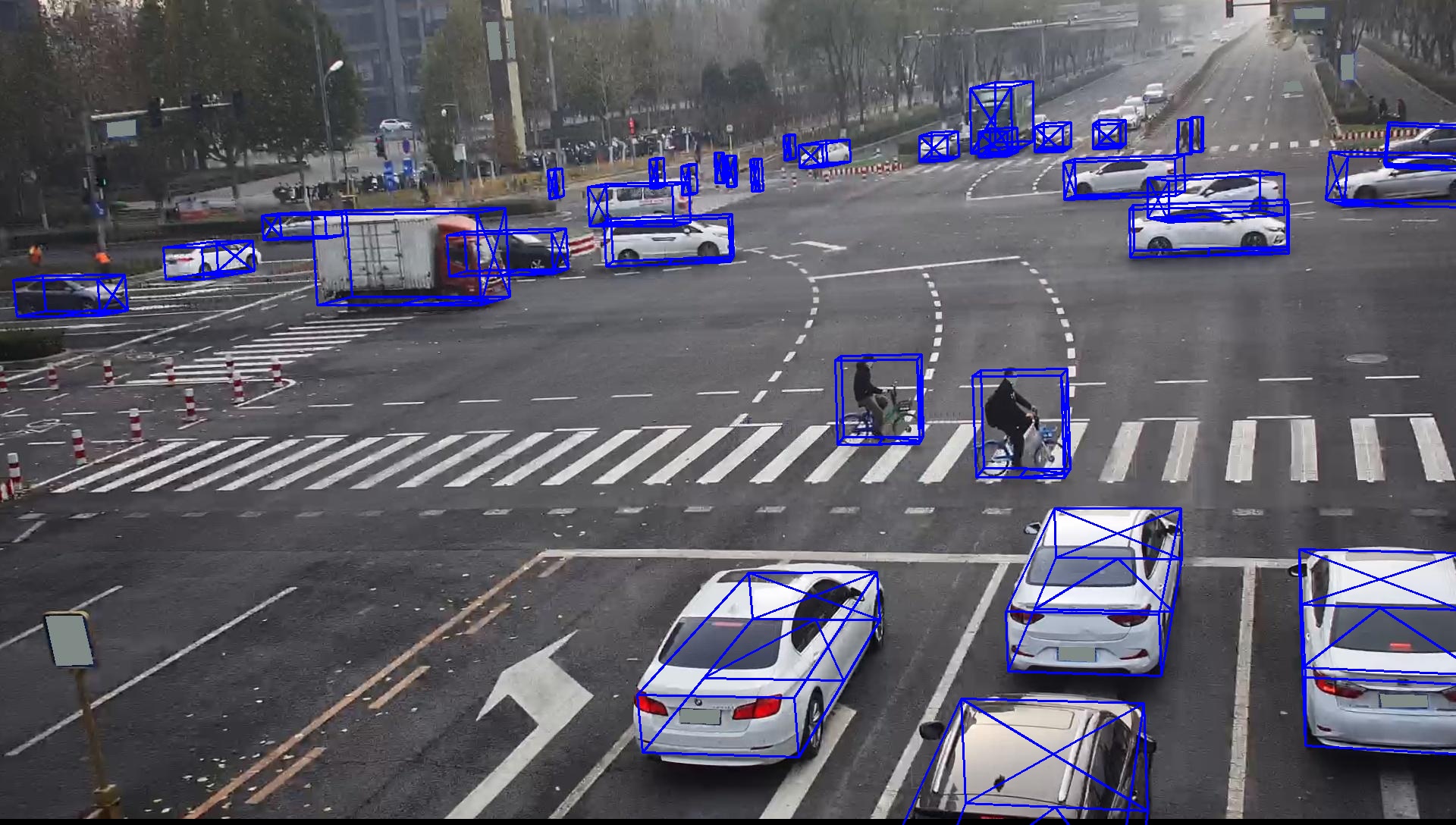}& 
\includegraphics[height=75pt]{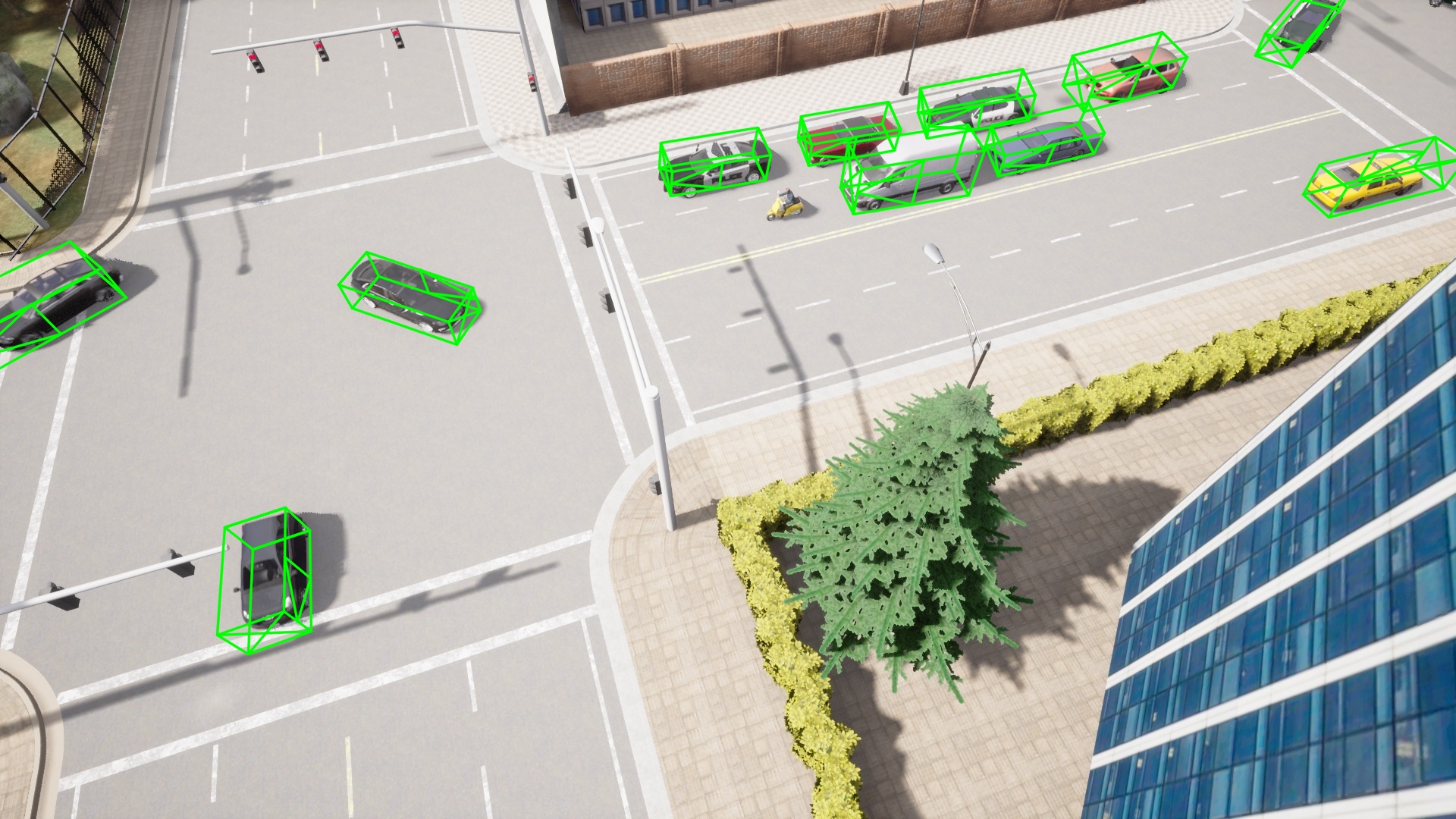} &
\includegraphics[height=75pt]{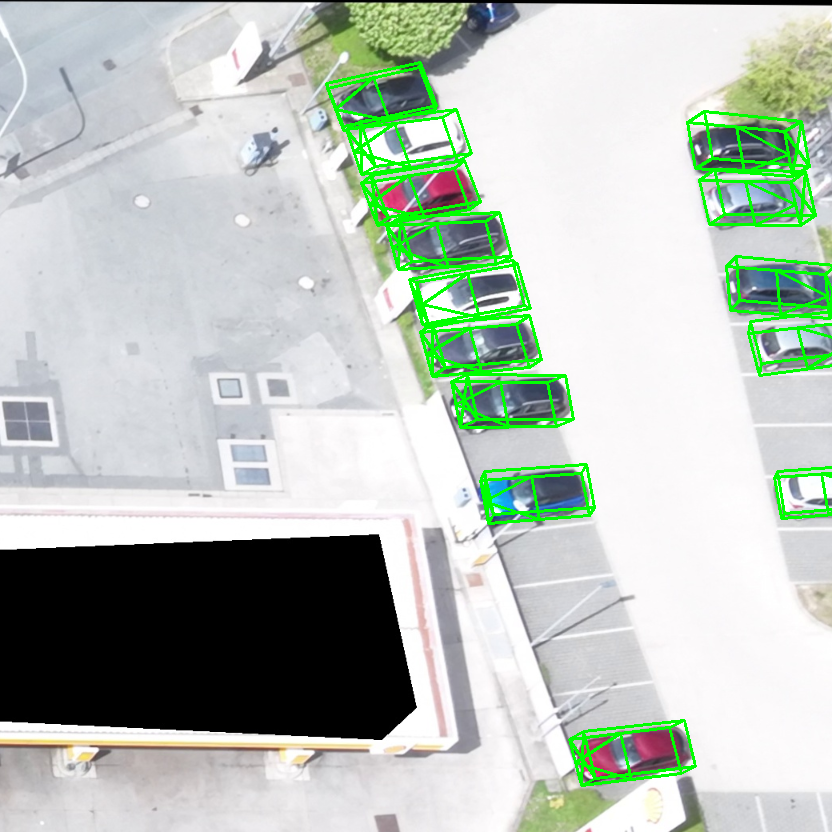}
\\

\end{tabular}
\caption{\textbf{Qualitative Results. } We compare our baseline MonoCon \cite{monocon} (first row) with \ourmethod (second row) across three datasets:
%For Rope3D \cite{Rope3D} and MonoCon \cite{monocon} we extend the single angle into a full 3D rotation using the provided ground normal. 
Rope3D \cite{Rope3D} (first column), CDrone (second column) and our in-house drone dataset (third column). We mask the building for privacy reasons. Our approach predicts more accurate dimensions and rotations. It has higher recall while minimizing false positives.}
\label{fig:main_qual_results}
\end{figure}

In \cref{fig:qual_results}, we qualitatively compare MonoCon \cite{monocon} with \ourmethod across available perspectives.
For Rope3D \cite{Rope3D} \groundmix leads to improved recall, including detecting distant and occluded objects. For CDrone, we extend the predictions of MonoCon \cite{monocon} to a full rotation matrix but encounter challenges in accurately predicting object rotation, potentially due to the limited viewpoint variety during training time. In the case of real-world drone data, we extend our baseline to predict $SO(3)$ rotation and to utilize virtual depth. As Rope3D, \ourmethod manages to detect more objects while also having fewer false positives. Likewise, we also observe more accurate predictions of the object dimensions and rotation.

\section{Conclusion}

Our work extends evaluation of monocular 3D detection with a challenging testbed comprising multi-perspective traffic scenes. 
As a novel contribution, we introduced \cdrone, a synthetic dataset of drone views.
We also designed an augmentation pipeline \groundmix, which improves a one-stage approach to operate consistently well on multiple datasets. 

Nevertheless, the detection accuracy of current state-of-the-art 3D detectors is generally insufficient for their real-world deployment.
We hope that this observation catalyzes the development of 3D detection models, capable of operating robustly across diverse camera setups.

\inparagraph{Limitations}
First, we acknowledge the synthetic nature of the proposed dataset \cdrone, which nevertheless possesses unique properties and challenges.
Second, our focus has been on the traffic setting, where we assume accurate ground plane fitting to the road is appropriate.
This assumption may be violated on uneven terrains and indoor scenes. 
Lastly, we experimented with a joint training setup across multiple perspectives, but did not observe a benefit, especially if the camera views vary significantly. This challenge is an interesting direction for future work towards unified 3D detection across all camera perspectives.

\footnotesize{
\inparagraph{Acknowledgements}
This work is a result of the joint research project STADT:up. The project is supported by the German Federal Ministry for Economic Affairs and Climate Action (BMWK), based on a decision of the German Bundestag. The author is solely responsible for the content of this publication. This work was also supported by the ERC Advanced Grant SIMULACRON. Additionally, the author would like to thank Zhuolun Zhou for his valuable technical contributions, particularly in the conversion of single-angle predictions to $SO(3)$ rotations.
}

% TODO: Replace with your title
\title{CARLA Drone: Monocular 3D Object Detection from a Different Perspective\\[2mm] \large -- Supplemental Material --} 
% You can use \thanks for acknowledgment. Do not add any acknowledgment to the draft 
% version that is used for the review process.  
%\title{Title\thanks{XXX}}
\titlerunning{CARLA Drone}

\ifreview
	% ANONYMOUS SUBMISSION FOR REVIEW
	% DO NOT MODIFY these for the draft version that is used for the review process.
	\titlerunning{GCPR 2024 Submission \SubNumber{}. CONFIDENTIAL REVIEW COPY.}
	\authorrunning{GCPR 2024 Submission \SubNumber{}. CONFIDENTIAL REVIEW COPY.}
	\author{GCPR 2024 - \GCPRTrack{}}
	\institute{Paper ID \SubNumber}
\else

	\author{Johannes Meier\inst{1,2,3} \quad
	Luca Scalerandi\inst{1,2} \quad
	Oussema Dhaouadi\inst{1,2,3} \\ 
        Jacques Kaiser\inst{1} \quad
        Nikita Araslanov\inst{2,3} \quad
        Daniel Cremers\inst{2,3}}
	
	\authorrunning{Meier et al.}
	
	\institute{$^1$\href{https://www.deepscenario.com}{DeepScenario}  \quad $^2$TU Munich \quad $^3$Munich Center for Machine Learning}
\fi

\maketitle              % typeset the header of the contribution

\appendix

\section{\ourmethod: Further details}

We offer additional material on \ourmethod. We first compare our method with existing approaches in \cref{sec:method_comparison}, next we detail the model architecture and contrast it with our baseline, MonoCon \cite{monocon} in \cref{sec:arch}. \cref{sec:groundmix_details} elaborates on the implementation details of our patch-pasting approach.
Finally, we provide further detail on the 2D/3D consistent rotation augmentation in \cref{sec:2d3drot}.

\subsection{Conceptual comparison to previous work}\label{sec:method_comparison}

%\begin{wraptable}[18]{r}{1.0\textwidth}
\begin{table*}[t]
    \caption{\textbf{Monocular 3D detection methods in comparison. } We show, whether car view (C), traffic camera view (T) and drone view (D) fulfill the assumptions of the corresponding methods. }
    %\centering
    \scriptsize
    \begin{tabularx}{\linewidth}{
        @{}X c@{\hspace{0.7em}}
             c@{\hspace{0.7em}}
             c@{\hspace{2em}}
             l}
        
      \toprule
      Method & C & T & D & Architecture \\ \midrule
      
      MonoRCNN++ \cite{MonoRCNN++} & \cmark & & & Two-Stage \\
      Cube R-CNN \cite{omni3d} & \cmark & \cmark & \cmark & Two-Stage \\ \midrule

      GUPNet \cite{GUPNet} & \cmark & & & One-Stage \\
      MonoFlex \cite{Monoflex} & \cmark & & & One-Stage \\
      DEVIANT \cite{DEVIANT} & \cmark & & & One-Stage \\
      MonoDDE \cite{MonoDDE} & \cmark & &  & One-Stage \\
      M3D-RPN \cite{M3D-RPN} & \cmark & & & One-Stage \\
      MonoUNI \cite{MONOUNI} & \cmark & \cmark & & One-stage \\
      MonoGAE \cite{MonoGAE} & & \cmark & & One-Stage \\ \midrule
      BEVFormer \cite{BEVFormer} & \cmark & \cmark & & BEV \\
      BEVDepth \cite{bevdepth} & \cmark & \cmark & & BEV \\
      BEVHeight \cite{BEVHeight} & & \cmark & & BEV \\ 
      \midrule
      This work & \cmark & \cmark & \cmark & One-Stage \\
    \bottomrule
    \end{tabularx}
    \label{tab:methods}
\end{table*}
%\end{wraptable}

\cref{tab:methods} demonstrates that our method effectively supports car view, traffic view, and drone view while maintaining high speed, thanks to the one-stage approach. Among the other methods, only Cube R-CNN \cite{omni3d} also supports all three views. However, it is more unstable during training time \cite{uniMode} and experiences slower training times.

\subsection{Model architecture}\label{sec:arch}
\begin{figure}[b]
  \centering
  \includegraphics[width=1.0\textwidth]{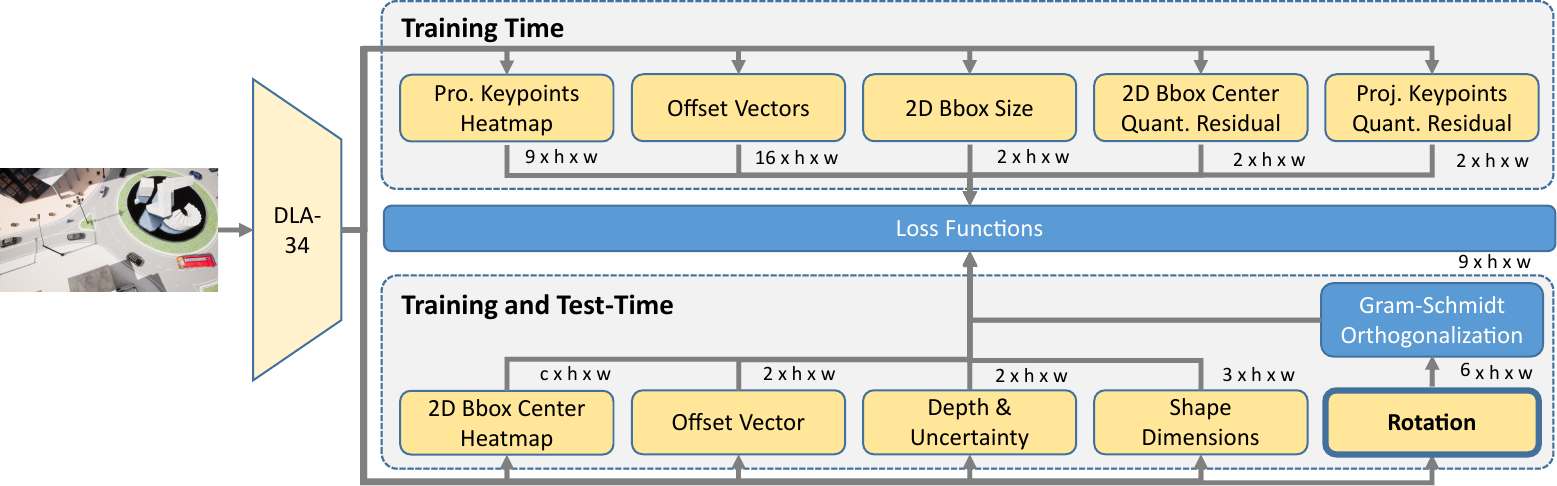}
  \caption{The MonoCon~\cite{monocon} architecture extended for \ourmethod.}\label{fig:model_architecture}
\end{figure}

In \cref{fig:model_architecture} we show the MonoCon architecture \cite{monocon} modified for \ourmethod. As in MonoCon~\cite{monocon}, we utilize auxiliary losses during the training time. 

MonoCon \cite{monocon} predicts 1DoF rotation around the $y$-axis of the camera.
\cref{fig:camera_coordinate_system} illustrates the camera's frame of reference.
We modify this head to enable prediction for the full orientation in $SO(3)$. The network predicts 6 parameters per object, which we then orthogonalize with the Gram-Schmidt method \cite{RotContinuity}.

Empirically, we observed that the keypoints predicted by MonoCon as an auxiliary task enhance accurate rotation estimation, which proved crucial for drone-view data. 
This further justifies our choice of using MonoCon as the baseline in our multi-perspective benchmark. 

\begin{figure}[t]
  \centering
  \begin{subfigure}[b]{0.13\textwidth}
    \includegraphics[width=\textwidth]{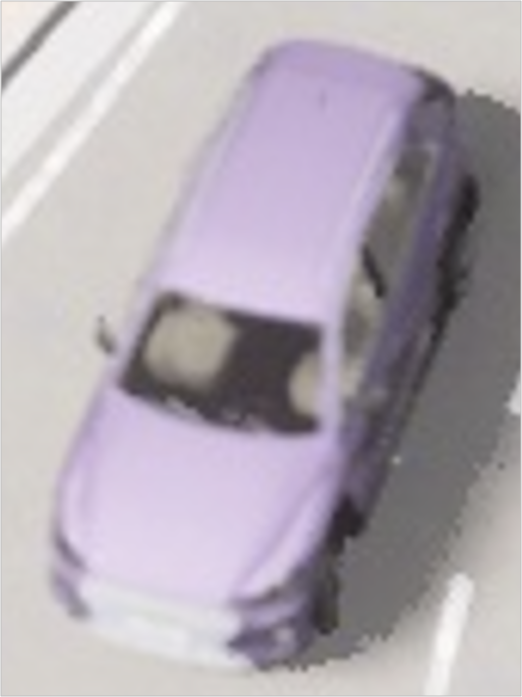}
    \caption{}\label{fig:groundmix1}
  \end{subfigure}\hspace{2em}
  \begin{subfigure}[b]{0.13\textwidth}
    \includegraphics[width=\textwidth]{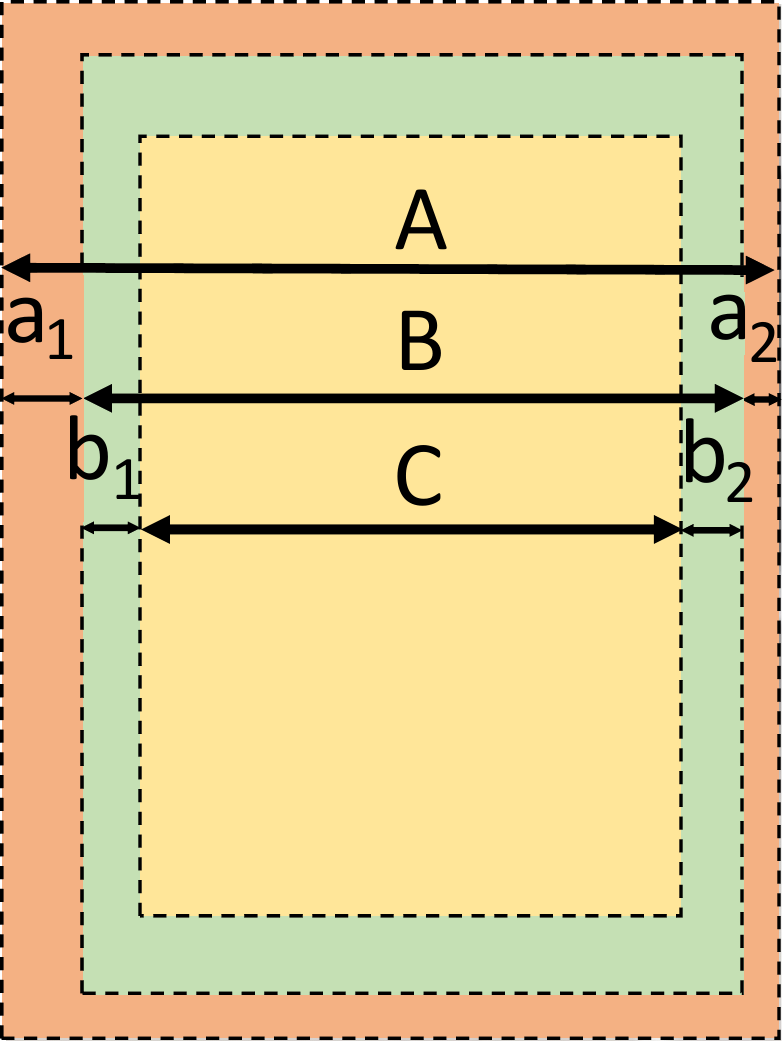}
    \caption{}\label{fig:groundmix2}
  \end{subfigure}\hspace{2em}
  \begin{subfigure}[b]{0.13\textwidth}
    \includegraphics[width=\textwidth]{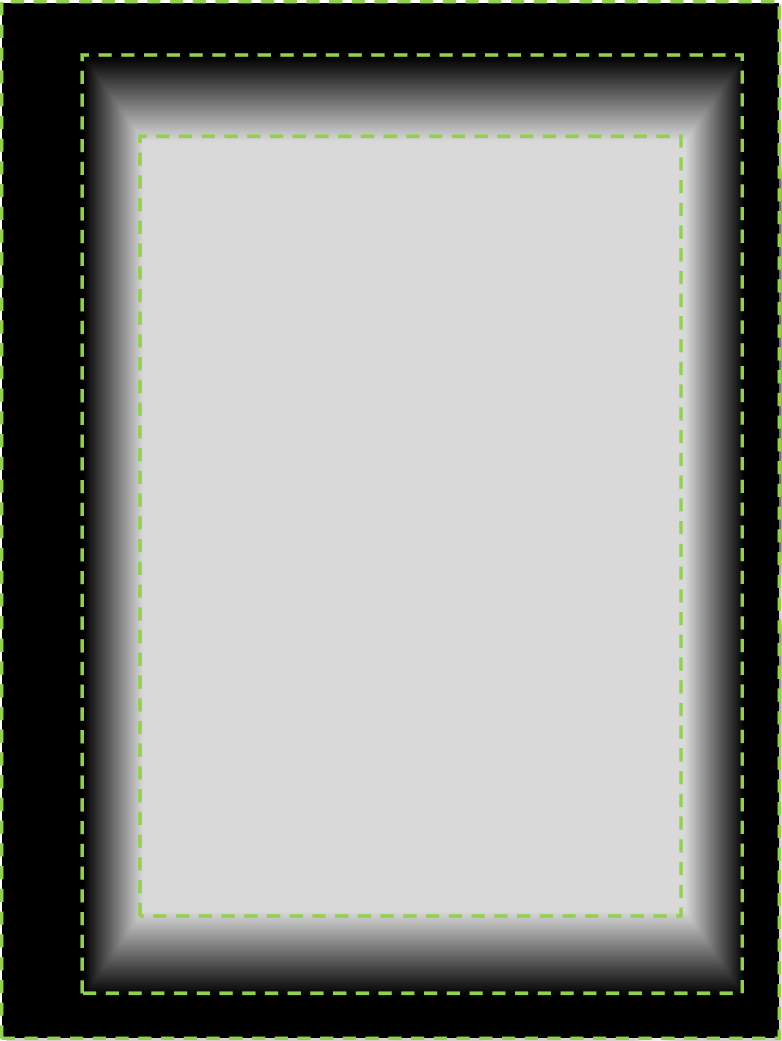}
    \caption{}\label{fig:groundmix3}
  \end{subfigure}\hspace{2em}
  \begin{subfigure}[b]{0.13\textwidth}
    \includegraphics[width=\textwidth]{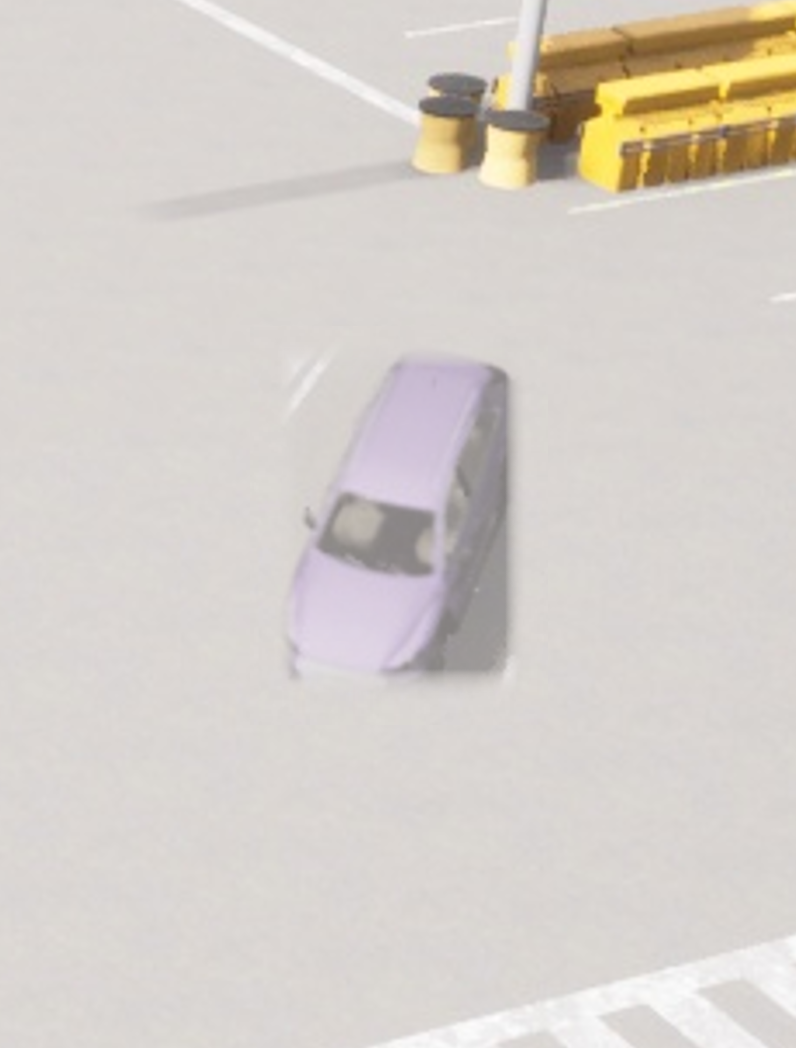}
    \caption{}\label{fig:groundmix4}
  \end{subfigure}\hspace{2em}
    \begin{subfigure}[b]{0.13\textwidth}
    \includegraphics[width=\textwidth]{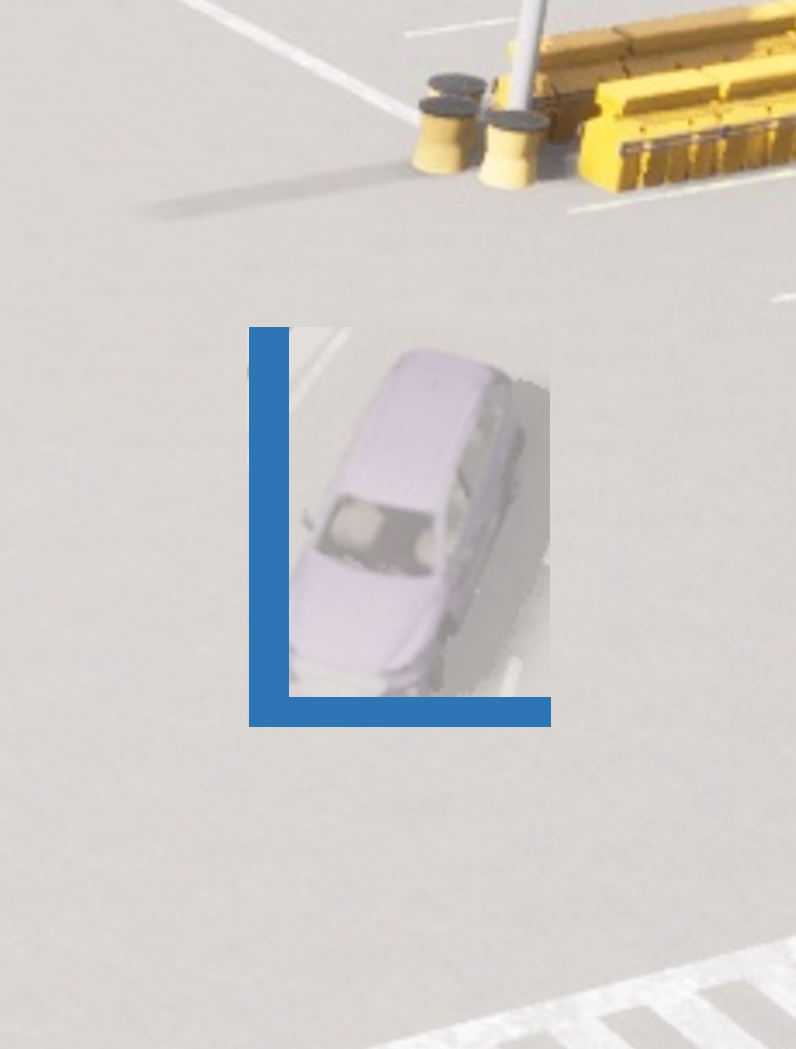}
    \caption{}\label{fig:groundmix5}
  \end{subfigure}
  
  \caption{Soft pasting in comparison to the pasting in Mix-Teaching \cite{mix-teaching}. \textbf{(a)} The patch to be pasted. \textbf{(b)} Sampling of the mask ($a_1, a_2 \in [0, 10\%\cdot A]$, $b_1, b_2 \in [0, 20\%\cdot B]$) \textbf{(c)} The corresponding mask, where the black denotes 0\% opacity and the white denotes 100\% opacity. \textbf{(d)} The result after pasting the object softly with the blending mask \textbf{(c)}. \textbf{(e)} The pasting as done in Mix-Teaching \cite{mix-teaching}. By contrast, our result \textbf{(d)} does not provide visual hints about the 2D bounding box.} 
\end{figure}

\subsection{\groundmix: Ground plane fitting}
\subsubsection{A. Fit ground plane, sample positions}
\label{sec:groundmix_details}
\groundmix requires a ground plane to sample new object locations.
We assume that an image has at least three objects located on the ground. Let $(x_1, y_1, z_1), ... (x_n, y_n, z_n)$ be the bottom center of their 3D bounding boxes.
We fit a ground plane $n_x x + n_y y + n_z z = d$ using least-squares estimation.  Note that the normal vector $(n_x, n_y, n_z)$ has unit length. Denote by $(\bar{x}, \bar{y}, \bar{z})$ the centroids and $X$ the normalized object locations, computed as:

\begin{equation}
     (\bar{x}, \bar{y}, \bar{z}) = \bigg(\frac{1}{n}\sum_i x_i, \frac{1}{n}\sum_i y_i, \frac{1}{n}\sum_i z_i \bigg)
\end{equation}

\begin{equation}
     X = \begin{bmatrix} x_1 - \bar{x} & & ... & & x_n - \bar{x} \\ y_1 - \bar{y} & & ... & & y_n - \bar{y} \\ z_1 - \bar{z} & & ... & & z_n - \bar{z} \end{bmatrix}
\end{equation}

\noindent The left-singular vector corresponding to the smallest singular value of matrix $X$ is the least-squares estimate of the plane's normal $(n_x, n_y, n_z)$.
Assuming that the centroid lies on the ground, we derive the remaining unknown $d$:
\begin{equation}
    d = n_x \bar{x} + n_y\bar{y} + n_z\bar{z}.
\end{equation}

\subsubsection{B. Hard mining / D. Patch buffer}
For hard-mining, we utilize the recent training loss as an indication of the object's difficulty. 
We initialize all difficulty measures with the maximum value at the beginning of training and update them after each gradient step. 
We include all losses except heatmap-based losses, as they are challenging to assign to a single object. 
One nuance with our approach concerns distant objects, which typically have a higher (depth) loss.
To prevent oversampling of these examples, we divide the target depth range into $b$ equally sized bins and randomly sample objects falling within each bin.
We use $b=6$ throughout our experiments. 

\subsubsection{C. Soft pasting} 
Our method \emph{seamlessly} integrates object patches into the target image, as depicted in \cref{fig:groundmix1,fig:groundmix2,fig:groundmix3,fig:groundmix4,fig:groundmix5}.
Naively pasting objects without smooth masking leads to a critical issue.
An abrupt transition between the pasted patch and the target image compromises the natural spatial consistency.
As a result, the detection network may learn to leverage these artificially created cues and fail to generalize. 
Therefore, we employ a three-stage masking strategy before pasting each object:
\textbf{Step 1:} We crop the original patch at the image boundary. To achieve this, we uniformly sample between 0\% to 10\% of the original dimension to remove along each side (highlighted in orange in \cref{fig:groundmix2}). This method ensures that the network cannot recover 3D information from the transition between the patch and the original image.
\textbf{Step 2:} Within this cropped area, we utilize the subsequent 0\% to 20\% of the area on each side to linearly increase visibility through MixUp \cite{mixup} (illustrated in green in \cref{fig:groundmix2}). This enhances image consistency, and is particularly useful when pasting objects from nighttime scenes into daytime images.
\textbf{Step 3:} Furthermore, we apply MixUp \cite{mixup} in the central area, with an opacity range of 80\% to 100\%, thereby introducing additional complexity to object detection (highlighted in yellow in \cref{fig:groundmix2}). 

When pasting objects, we consider the physical surrounding of a patch. To account for differences in focal length and the depth at the new target location, we scale object patches by $s$. % (see \cref{sec:soft_pasting}). 
For $s > s_\text{max}$ we refrain from pasting to avoid blurry objects. We set an upper bound of $s_\text{max} = 2.0$. 
We also exclude patches with intruding objects, using $35\%$ as the cut-off threshold for the overlap with their respective 2D bounding box.

\subsection{2D-3D consistent rotation augmentation}\label{sec:2d3drot}

\begin{figure}[t]
\begin{minipage}[c]{0.24\linewidth}
\includegraphics[width=0.9\linewidth]{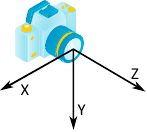} 
\vspace{10pt}
\caption{Camera coordinate system.}
\label{fig:camera_coordinate_system}
\end{minipage}
\hfill
\begin{minipage}[c]{0.80\linewidth}
\includegraphics[width=\linewidth]{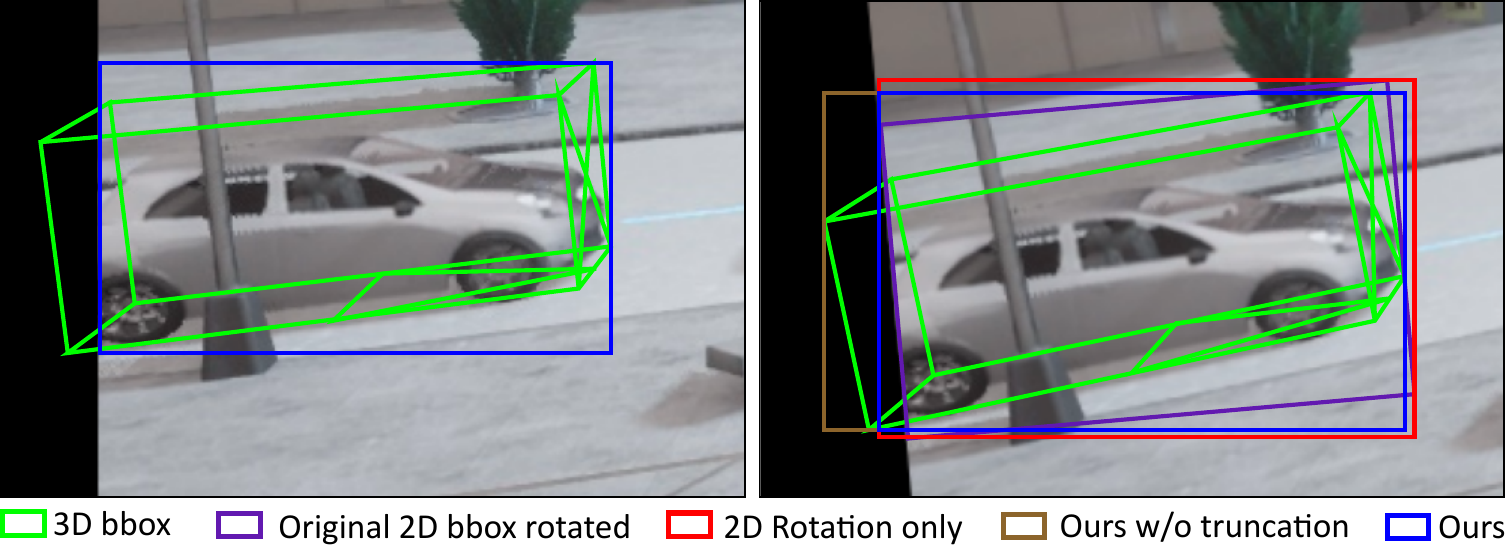}
\caption{Applying 5° rotation augmentation on an object at the image boundary. The left and right image show the object before and after rotation. In contrast to other possible approaches, our 2D bounding box remains tight and 2D/3D consistent.}
\label{fig:rot_example}
\end{minipage}%
\end{figure}

Unlike traffic view and car view, rotation of the camera along its optical axis are natural in a drone dataset. While training, we apply rotation augmentation on the images from the CDrone dataset. In this section, we explain how the 3D labels are transformed accordingly. 

The 2D image rotation can be seen as a rotation of the camera in 3D. In other words, rotation of an image by an angle $\phi$ is equivalent to a in-plane rotation of the camera by $-\phi$. Let $R \in SO(3)$ be the egocentric rotation matrix before applying the transformation.
Then, we obtain the rotation matrix $R_{\phi} \in SO(3)$ from additionally rotating by angle $\phi$: 
\begin{equation}
     R_{\phi} = \begin{bmatrix} cos(\phi) & -sin(\phi) & 0 \\ sin(\phi) & cos(\phi) & 0 \\ 0 & 0 & 1 \end{bmatrix} R.
\end{equation}

We also have to adjust the object center. For that, we first project the 3D object center onto the image. Next, we rotate the projected point in pixel space and unproject with the original depth to 3D space again. 

Illustrated in \cref{fig:rot_example}, this approach guarantees consistency between the 2D bounding box and the corresponding 3D bounding box.
Additionally, we demonstrate the impact of truncating the bounding box with the rotated original 2D bounding box in this example.
While this truncation is not necessary for maintaining 2D-3D consistency, it does result in a tighter bounding box, ensuring coverage over only the visible area.

\subsection{Implementation details}
We use DLA-34 \cite{DLA-34} as our backbone. The model is trained on a single GPU for 90 epochs, with a batch size of $8$. We employ AdamW \cite{AdamW} as our optimizer and a cosine learning rate scheduler. The learning rate and the weight decay are set to $2.25 \times 10^{-4}$ and $10^{-5}$, respectively. For drone data, rotation augmentation is performed with 50\% probability, while we do not use it for car and traffic view data. We apply MixUp \cite{mixup} and scale augmentation with $50\%$ probability on the training sample. To reduce the training time, we downscale the images by $50\%$ \wrt their original size in case of Waymo \cite{waymo} and Rope3D \cite{Rope3D}, and by $75\%$ in case of \cdrone. We paste up to 6 objects per image, as we observe performance saturation when pasting more than 4–6 objects. 

\section{Experiments}

\subsection{Waymo}

In \cref{tab:waymo_full} and \cref{tab:waymo_lvl1}, we provide the complete Waymo \cite{waymo} evaluation results from the main text, which was truncated due to space limitations. For MonoCon \cite{monocon} and \ourmethod we also report APH, which incorporates rotation information into AP.

\subsection{Recovering 3D rotation from a single angle}

Our drone evaluation on \cdrone requires $SO(3)$ rotation prediction.
However, MonoFlex \cite{Monoflex}, MonoCon \cite{monocon}, and DEVIANT \cite{DEVIANT} and related methods only support single-angle prediction. In these cases, we convert the single angle prediction to a $SO(3)$ rotation, using knowledge about the normal of the ground at this position and assuming the object lies flat on the ground.

We assume that all objects in the dataset have their $Z$-axis pointing upward, along the normal of the ground. We consider the single value predicted by the network to be the yaw angle around its normal, while roll and pitch are computed using the Euler angle convention to ensure the object lies flat on the ground. 

\subsection{In-house real-world drone data}
For drone data captured at altitudes exceeding 100 meters, 3D bounding box overlaps are infrequent, rendering AP3D an unsuitable metric. Alternatives like the NuScenes metric \cite{nuScenes} or mean average error either necessitate precise depth prediction or fail to account for the confidence score. Consequently, we propose evaluating the accuracy of predicted 3D bounding boxes separately for depth ($AP_{Depth}$) and the other 3D properties ($AP_{3DP}$), which include length, width, height, projected center, and $SO(3)$ rotation. A prediction qualifies as accurate for $AP_{Depth}^x$ if it overlaps with a ground truth object by more than 70\% in 2D and the depth error is within $x$ meters. The $AP_{Depth}$ is then computed as:
$$AP_{Depth} = \frac{1}{20} \sum_{x = 1, 2, ..., 20}AP_{Depth}^x.$$
For $AP_{3DP}$, we first match all predictions with ground truth objects in 2D. If there is an overlap above 70\%, we replace the predicted depth with the ground truth depth and then perform a standard $AP_{3D}$ evaluation at an IoU threshold of 50\%. We utilize these metrics for an in-house real world drone dataset. In contrast, we use the simpler $AP_{3D}$ metric for CDrone as it involves lower altitudes.

\subsection{Qualitative results}
\begin{figure}[t]
\centering
\begin{tabular}{c c c}

\includegraphics[height=66pt]{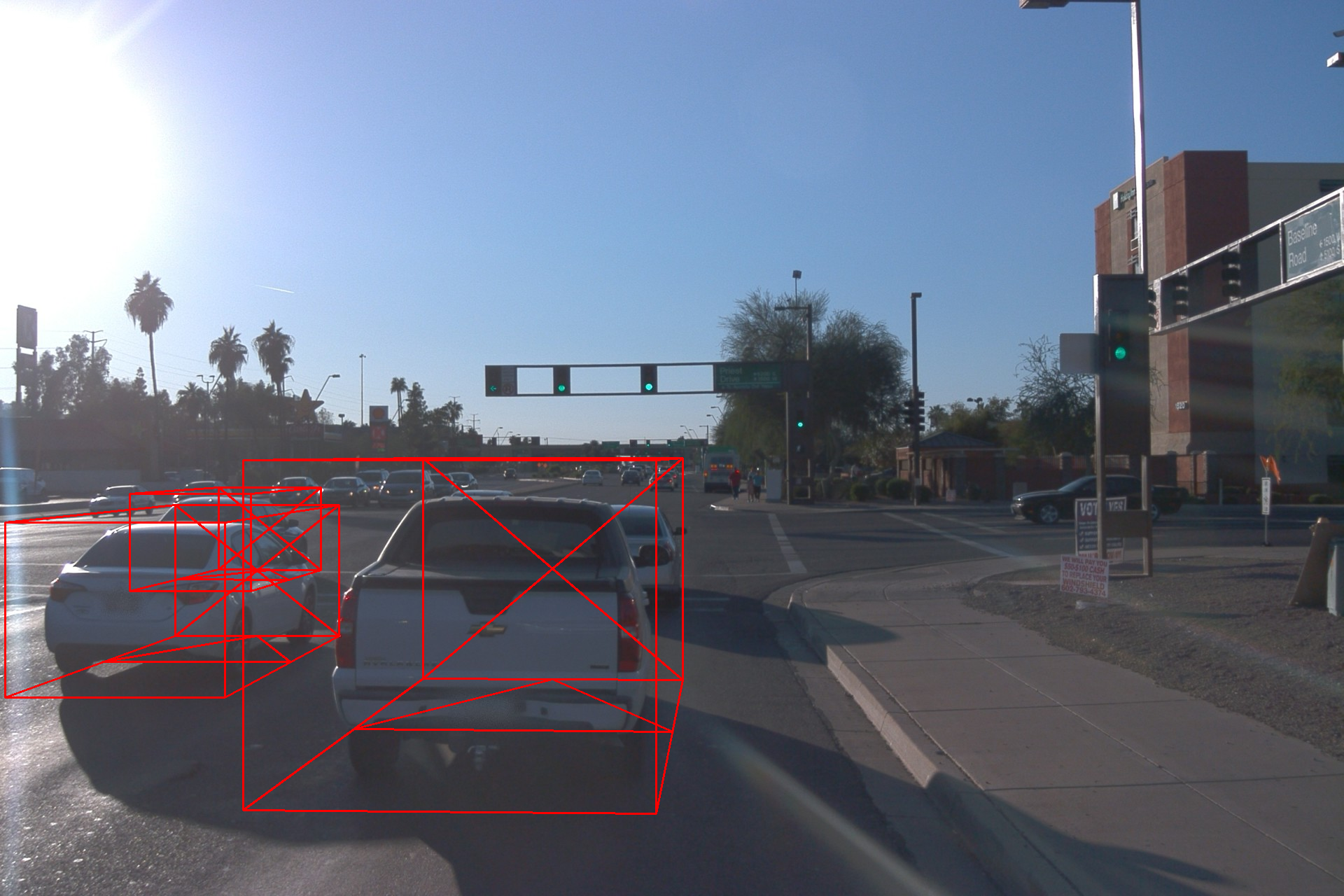} &
\includegraphics[height=66pt]{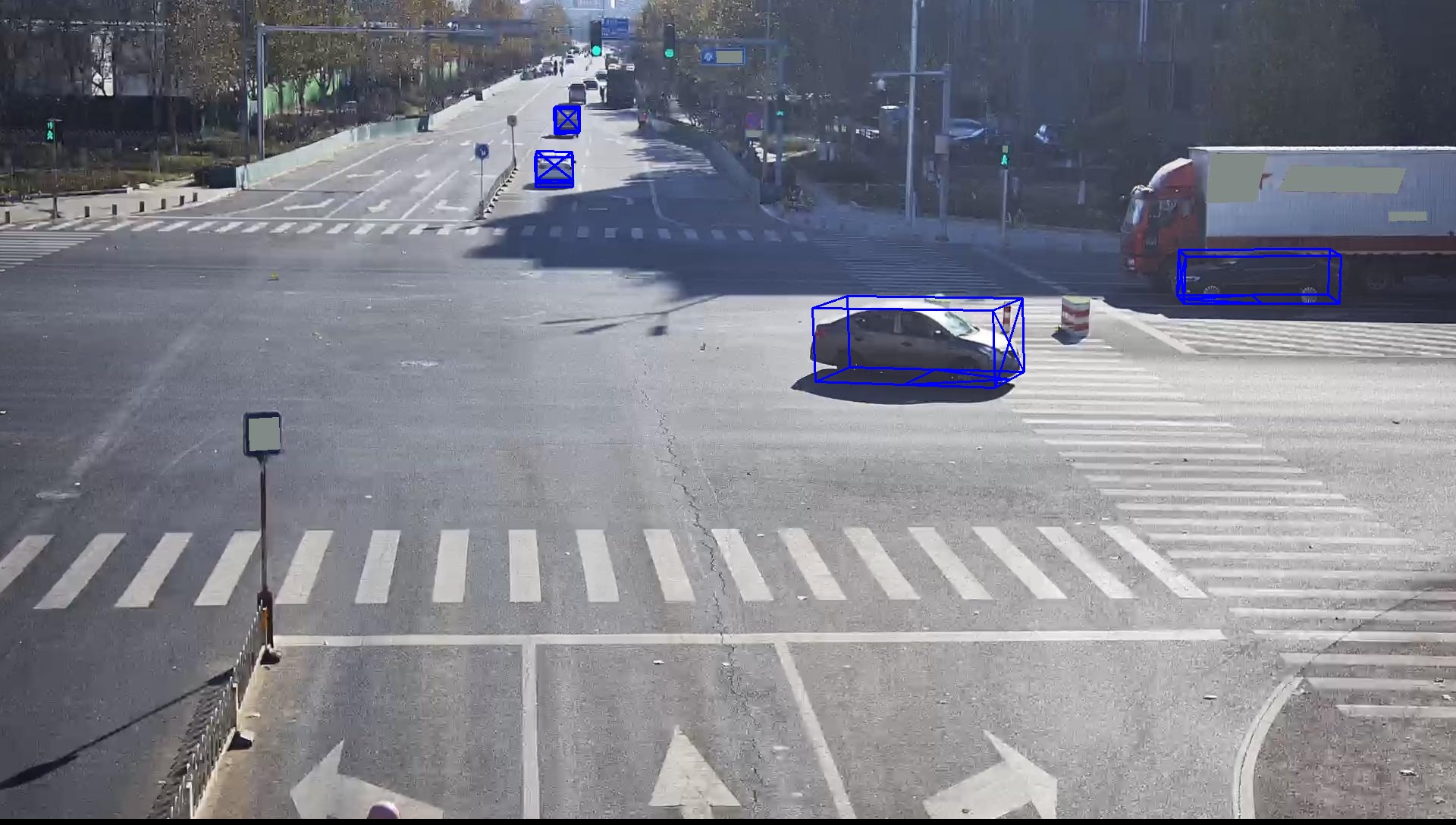} &
\includegraphics[height=66pt]{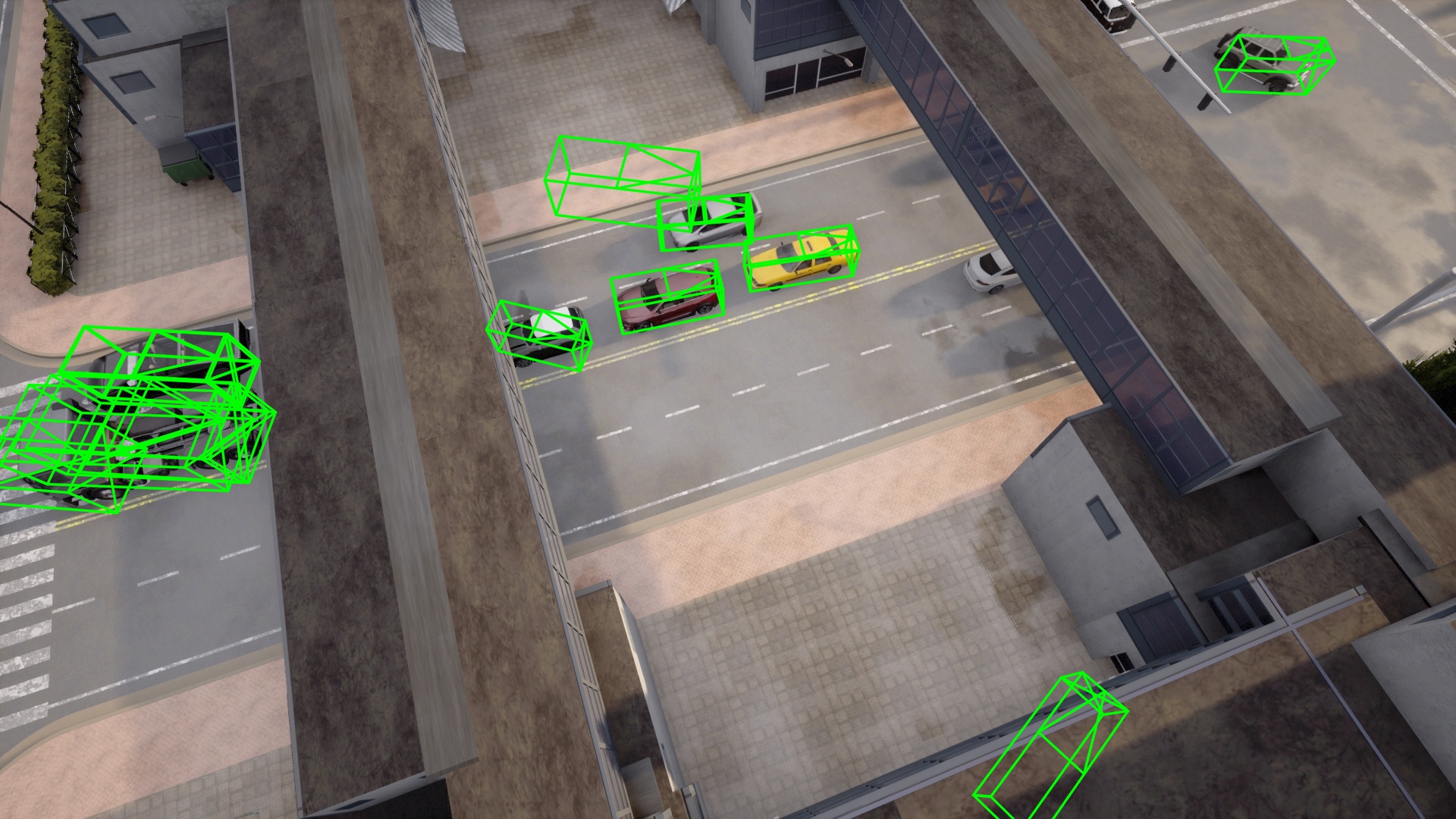} \\

\includegraphics[height=66pt]{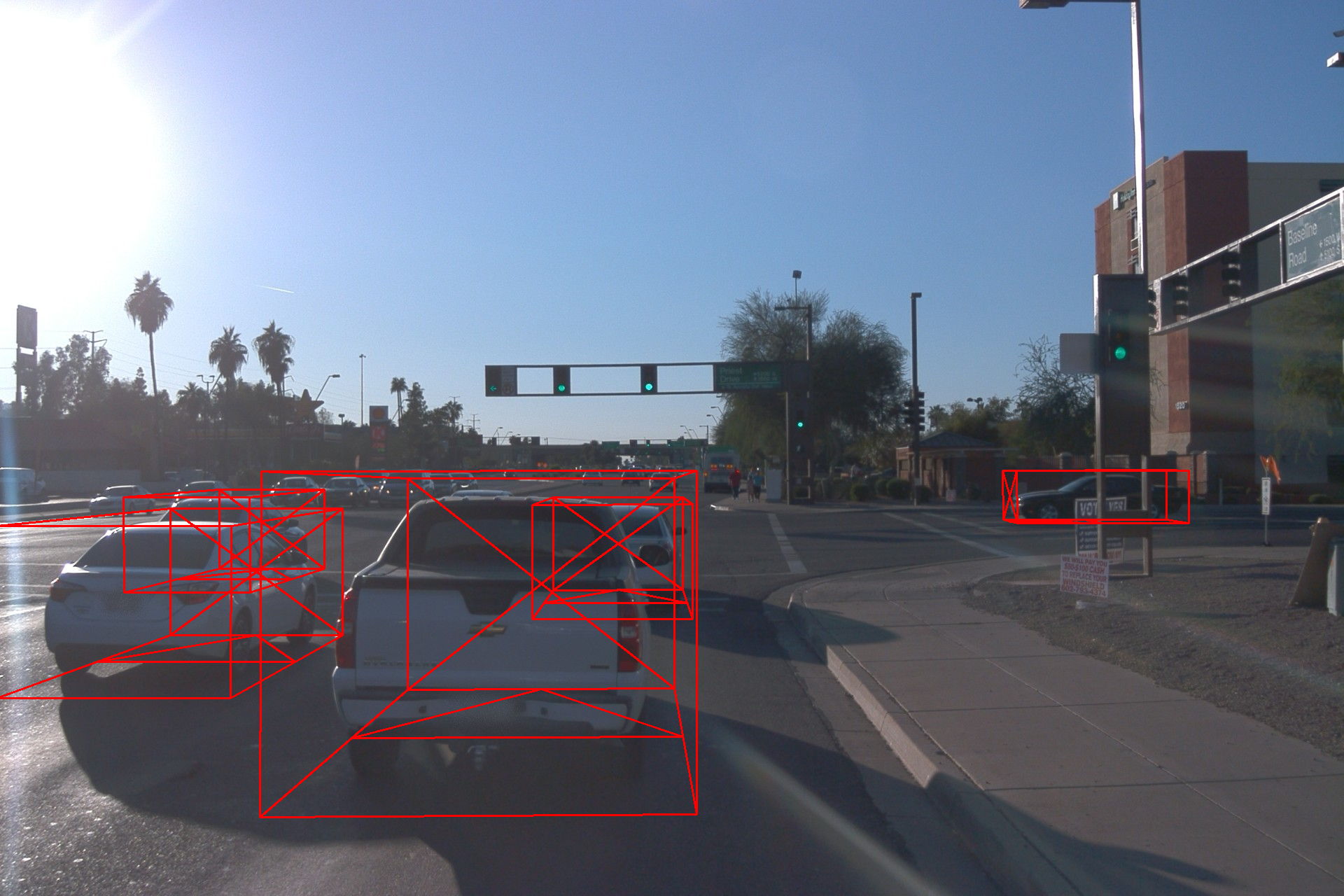} &
\includegraphics[height=66pt]{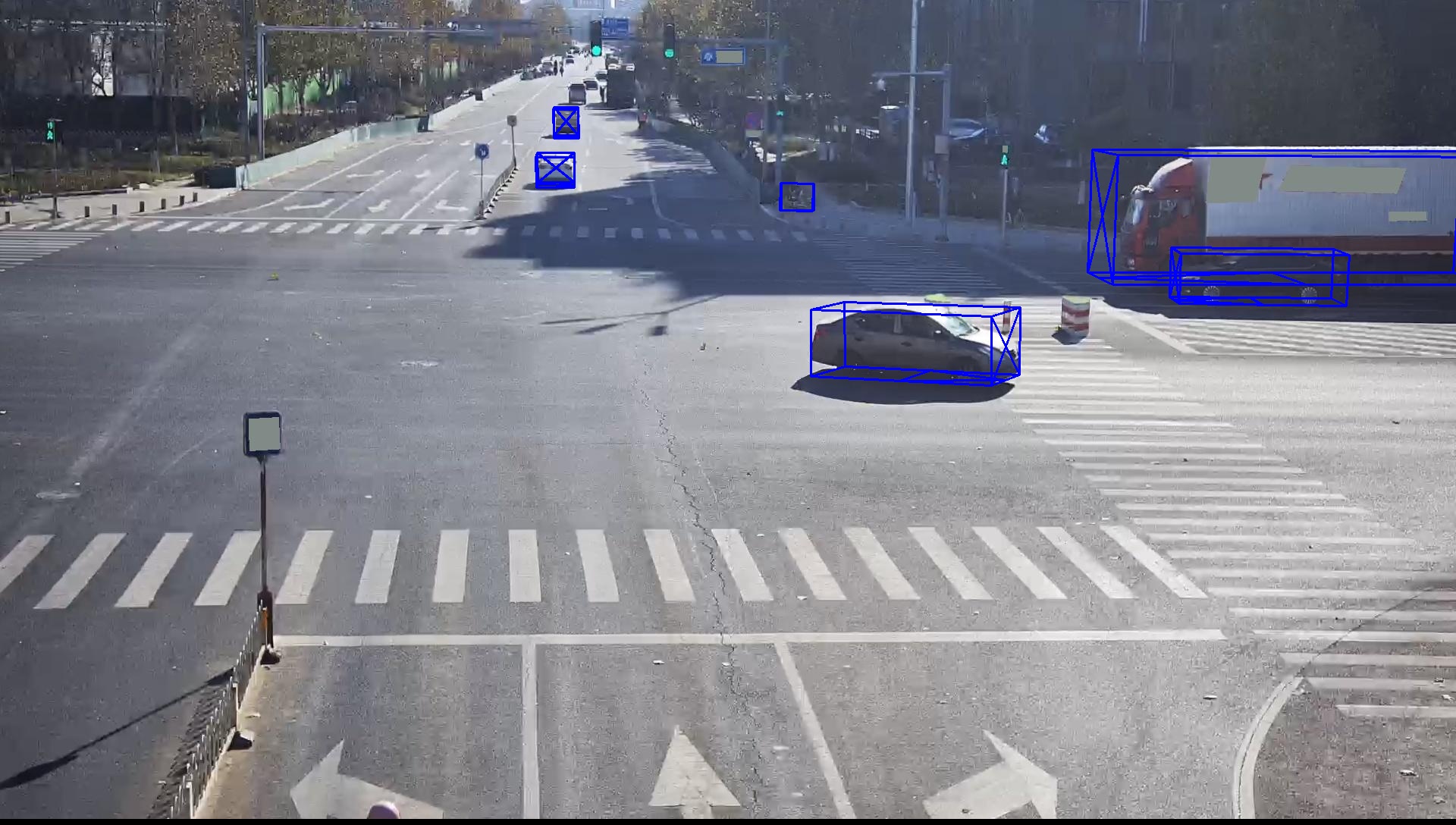}& 
\includegraphics[height=66pt]{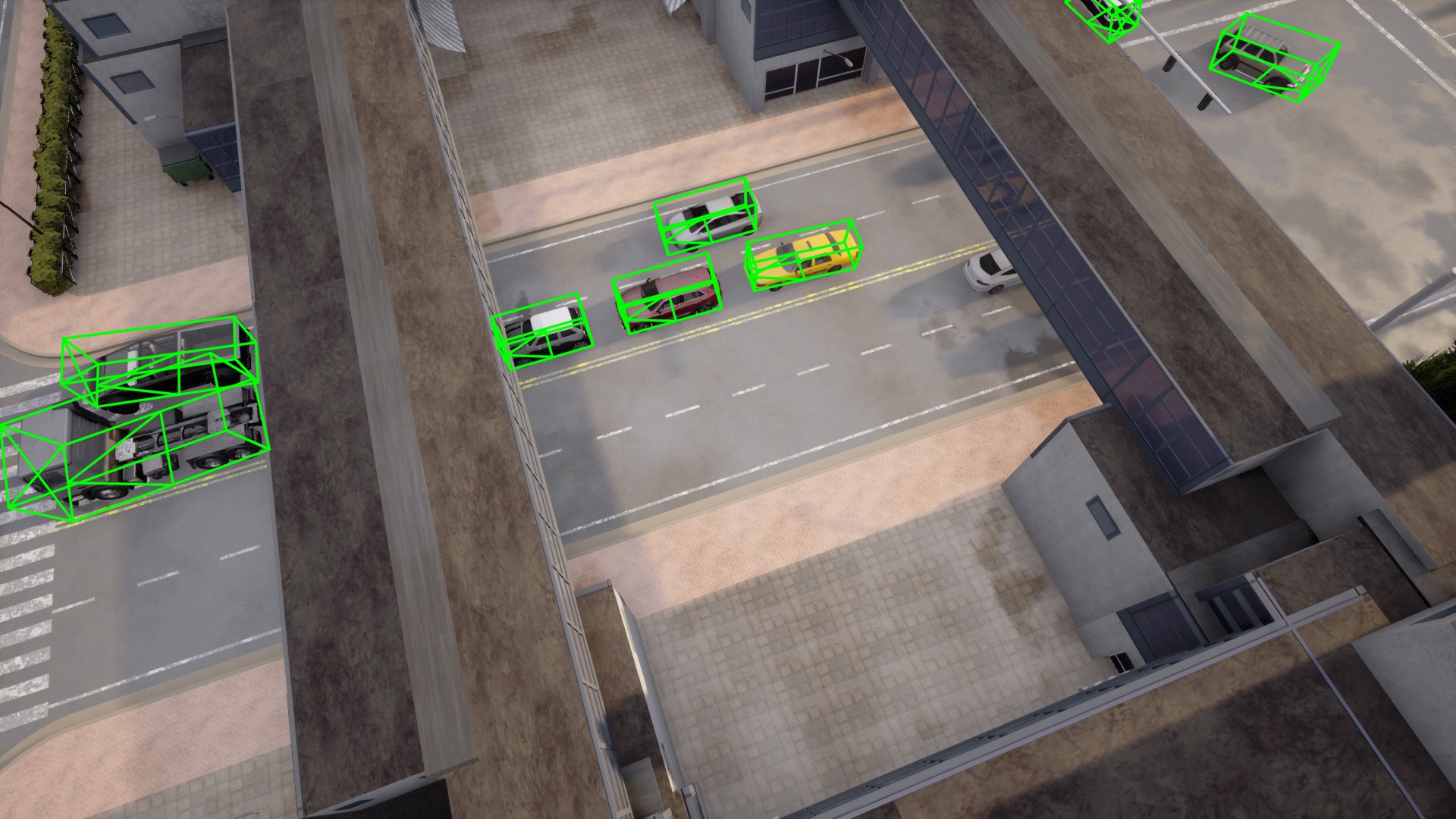} \\

\end{tabular}
\caption{\textbf{Qualitative Results. } We compare our baseline MonoCon \cite{monocon} (first row) with \ourmethod (second row). For MonoCon \cite{monocon} we extend the single angle into a full 3D rotation using the provided ground normal. The column indicates the dataset: Waymo \cite{waymo} (first column), Rope3D \cite{Rope3D} (second column), CDrone (third column).}
\label{fig:qual_results}
\end{figure}

\begin{figure}[t]
\centering
\begin{tabular}{c c c}

\includegraphics[height=115pt]{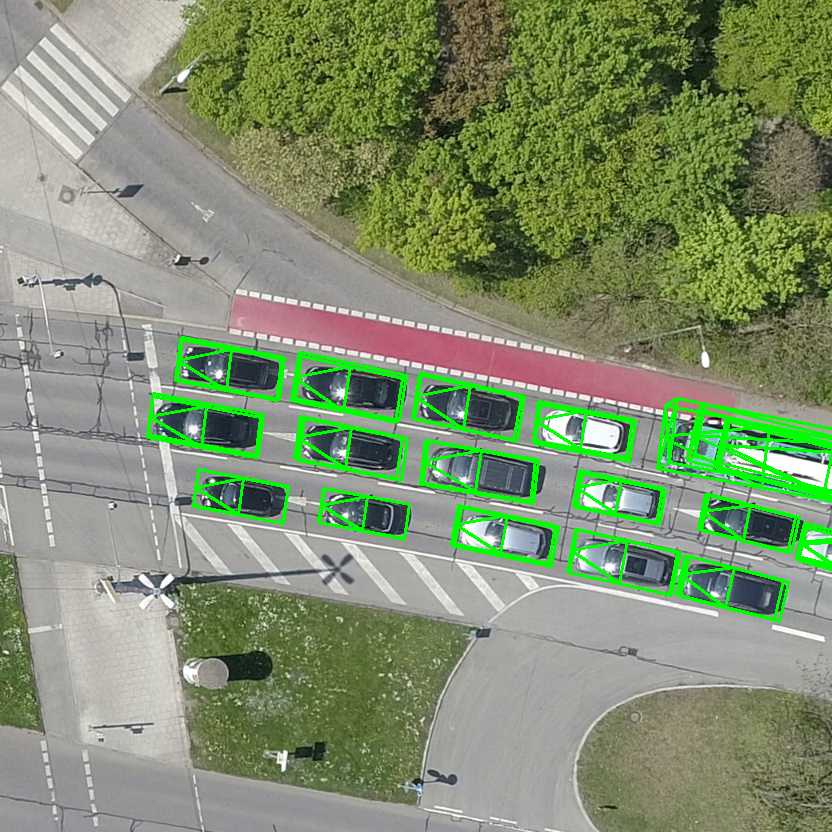} &
\includegraphics[height=115pt]{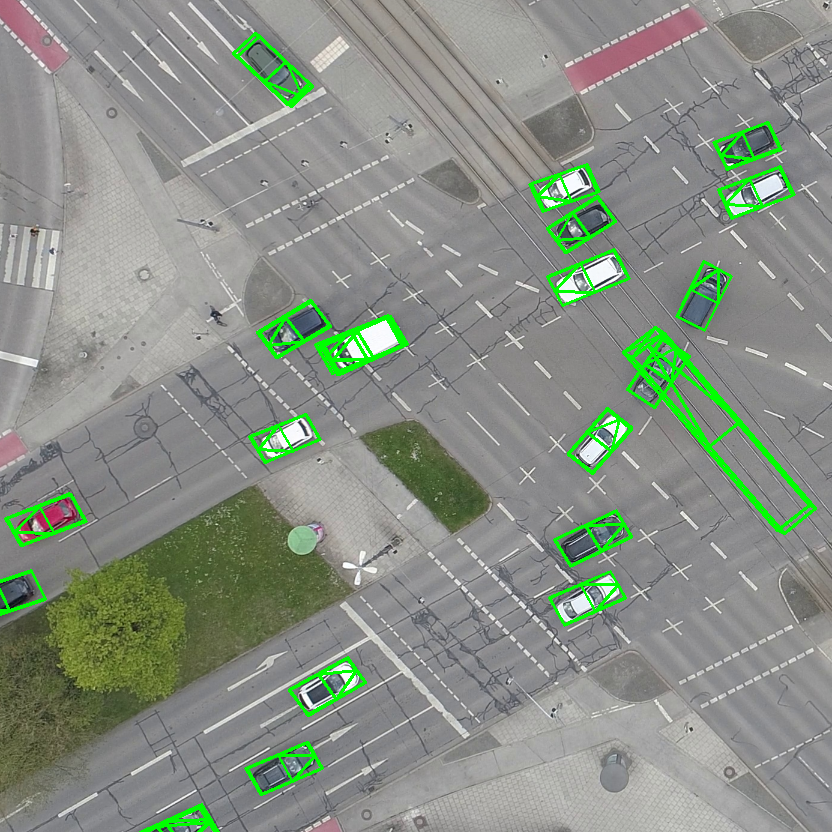} &
\includegraphics[height=115pt]{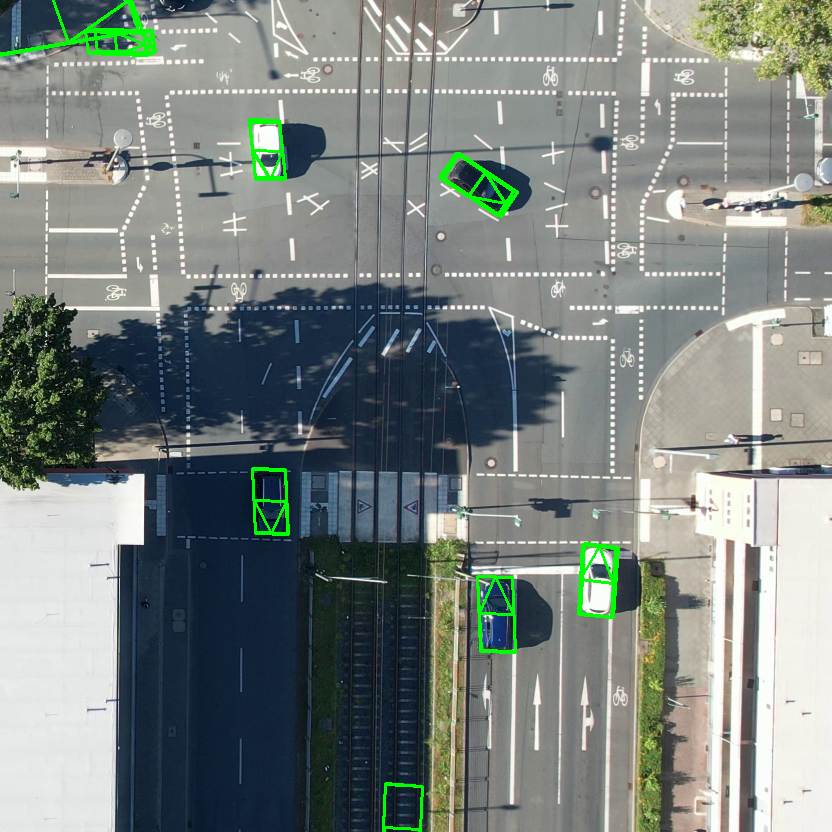}
\\

\includegraphics[height=115pt]{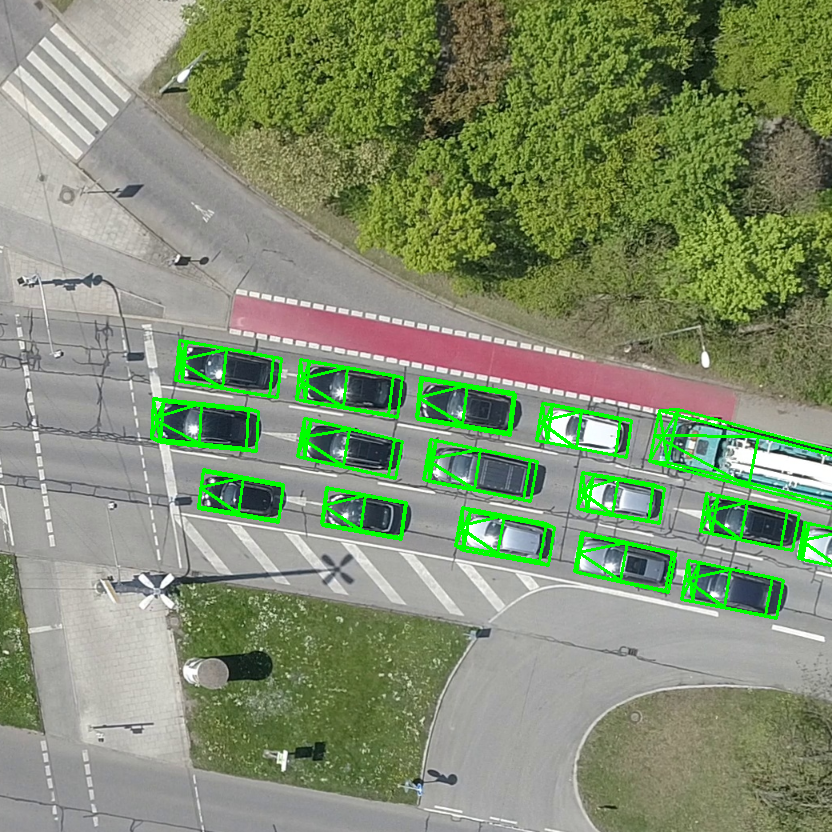}& 
\includegraphics[height=115pt]{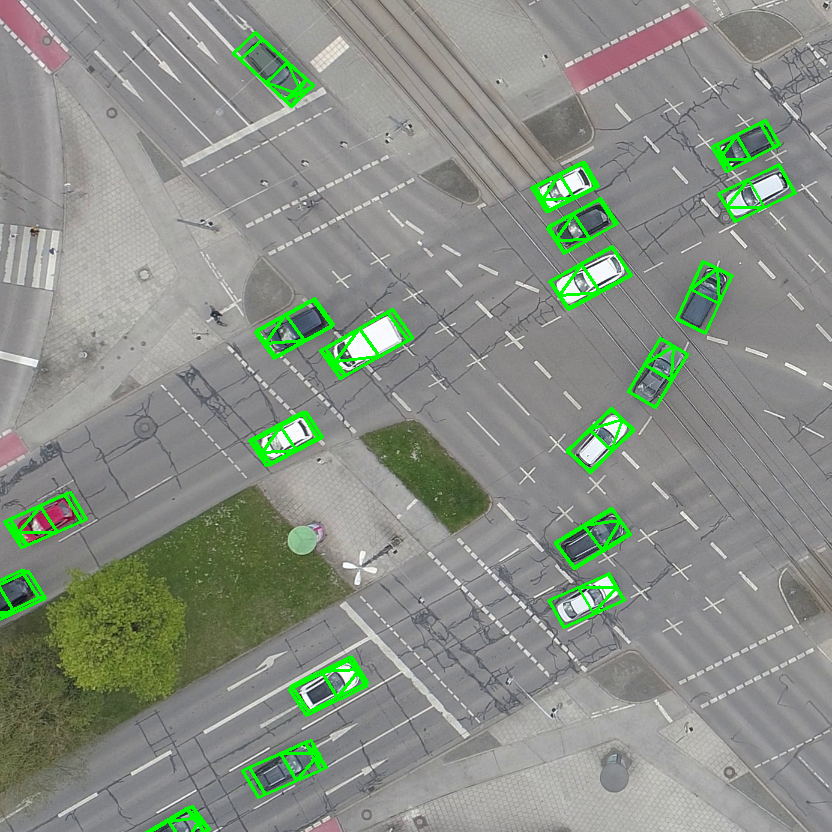} &
\includegraphics[height=115pt]{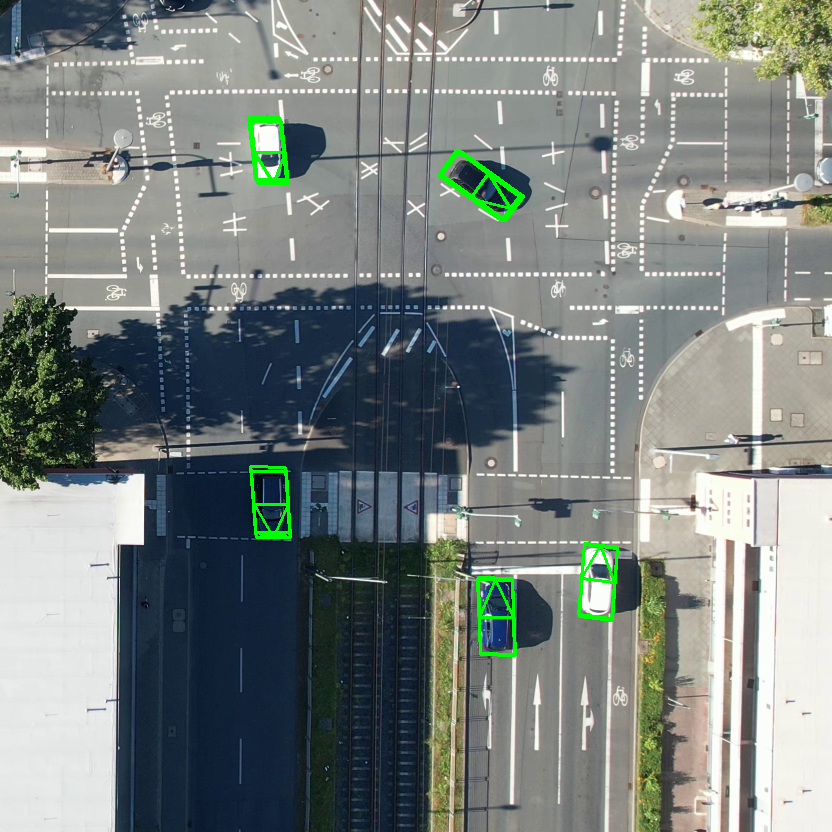}
\\

\end{tabular}
\caption{\textbf{Qualitative Results. } We compare our baseline MonoCon \cite{monocon} (first row) with \ourmethod (second row) on the in-house real drone dataset. Our approach predicts more accurate 3D box rotation and dimensions while leading to fewer false positives.}
\label{fig:qual_results_real_drone}
\end{figure}

In \cref{fig:qual_results} and \cref{fig:qual_results_real_drone} we show further qualitative results comparing \ourmethod to our baseline MonoCon \cite{monocon}. For real drone data, we extend MonoCon \cite{monocon} to additionally utilize virtual depth and predict $SO(3)$ rotation to ensure a fair comparison. In case of Waymo \cite{waymo} and Rope3D \cite{Rope3D} we observe more detections by using \ourmethod over MonoCon \cite{monocon}, particularly in case of occlusions. For both synthetic and real-world drone data, our method demonstrates more accurate rotation predictions, fewer false positives, and reduced double detections. We attribute this to our data augmentation pipeline that makes the model adapt more quickly to diverse viewpoints. 

\subsection{Ablation}

\begin{table}[t]
    \scriptsize
    \caption{\textbf{Ablation study of soft patch pasting.} The evaluation uses a subset of Rope3D \cite{Rope3D}. We study the effect of hard mining and compare to an alternative pasting strategy, Mix-Teaching \cite{mix-teaching}. We report $AP_{3D|R40}$ with $IoU = 0.5$.}
    \centering
    \begin{tabularx}{\linewidth}{
        @{}X @{\hspace{1em}}
        S[table-format=2.1]@{\hspace{1em}}
        S[table-format=2.1]
        }
      \toprule
      Method & Car & {Big Vehicle} \\
      \midrule
      \ourmethod & 44.1 & 28.4 \\
      w/o hard mining & 42.8 & 26.9 \\
      w/ Mix-Teaching \cite{mix-teaching} & 42.6 & 27.0 \\
      \bottomrule
      \end{tabularx}

    \label{tab:supp_ablation}
\end{table}

In \cref{tab:supp_ablation}, we present further ablation results with \groundmix. We apply the same training and validation splits as in the main text. Initially, we replace the hard mining strategy with uniform object sampling. This leads to a decline for the ``car'' category from $44.1$ to $42.8$ AP; the ``big vehicle'' scores decrease from $28.4$ to $26.9$. We hypothesize that hard mining benefits ``big vehicles'' more significantly due to their underrepresented nature in the dataset. Next, we evaluate the importance of pasting objects at our inferred ground plane. As a simple alternative, we apply the Mix-Teaching~\cite{mix-teaching} strategy of pasting objects at the same 2D bounding box locations as in the source image. However, changes in the camera altitude or angle can introduce inconsistencies within the image. Empirically, we observe decreased accuracy for the ``car'' class from $44.1$ to $42.6$ AP, and a reduction in ``big vehicle'' accuracy from $28.4$ to $27.0$ AP. This underscores the importance of pasting objects at physically plausible, 3D-consistent locations.
It further suggests that the scale augmentation integrated into \groundmix provides additional benefits beyond the existing image-level scale augmentation.

\begin{table*}[t]
    \caption{\textbf{3D detection accuracy on Waymo (val).} We use the data splits proposed by Reading \etal~\cite{waymo} for training and validation, and report the accuracy for the vehicle class across various distance thresholds. \ourmethod improves over the baseline and can operate across all camera views in our benchmark: the car view (C), the traffic camera view (T) and the drone view (D).}
    %$^*$DEVIANT\cite{DEVIANT} uses GUPNet \cite{GUPNet}, which is designed for car view \cite{MONOUNI}. 
    \setlength \scriptsize
    \centering
    \medskip
    %\resizebox{1.0\textwidth}{!}
    \begin{tabularx}{\linewidth}{
        @{}c  c  X  ccc @{\hspace{1em}}
        *{2}{S[table-format=2.2]@{\hspace{0.5em}}}
        S[table-format=2.2]@{\hspace{0.5em}}
        S[table-format=1.2]@{\hspace{1em}}
        *{2}{S[table-format=2.2]@{\hspace{0.5em}}} 
        *{1}{S[table-format=1.2]@{\hspace{0.5em}}} 
        S[table-format=1.2]}
        
      \toprule
      \multirow{2}{*}{IoU} & \multirow{2}{*}{Level} & \multirow{2}{*}{Method} & \multirow{2}{*}{C} & \multirow{2}{*}{T} & \multirow{2}{*}{D} & \multicolumn{4}{c}{$AP_{3D} \uparrow$} & \multicolumn{4}{c}{$APH_{3D} \uparrow$} \\
      \cmidrule(lr){7-10} \cmidrule(lr){11-14}
      & & & & & & {All} & {0-30} & {30-50} & {50-$\infty$} & {All} & {0-30} & {30-50} & {50-$\infty$} \\ \midrule
      
      \multirow{2}{*}{0.7} & \multirow{2}{*}{1} & MonoCon \cite{monocon,choi2023depthdiscriminative} & \cmark & \cmark &  & 2.30 & 6.66 & 0.67 & 0.02 & 2.29 & 6.62 & 0.66 & 0.02 \\
      & & \ourmethod (Ours) & \cmark & \cmark & \cmark & \textbf{3.10} & \textbf{8.94} & \textbf{1.17} & \textbf{0.06} & \textbf{3.08} & \textbf{8.88} & \textbf{1.16} & \textbf{0.06} \\ \midrule

      \multirow{2}{*}{0.7} & \multirow{2}{*}{2} & MonoCon \cite{monocon,choi2023depthdiscriminative}  & \cmark & \cmark &  & 2.16 & 6.64 & 0.64 & 0.02 & 2.15 & 6.59 & 0.64 & 0.02 \\
       & & \ourmethod (Ours) & \cmark & \cmark & \cmark & \textbf{2.73} & \textbf{8.81} & \textbf{1.06} & \textbf{0.04} & \textbf{2.71} & \textbf{8.75} & \textbf{1.05} & \textbf{0.04} \\  \midrule

      \multirow{2}{*}{0.5} & \multirow{2}{*}{1} & MonoCon \cite{monocon,choi2023depthdiscriminative}  & \cmark & \cmark &  &  10.07 & 27.47 & 3.84 & 0.17 & 9.99 & 27.26 & 3.81 & 0.16\\
       & & \ourmethod (Ours) & \cmark & \cmark & \cmark &  \textbf{11.89} & \textbf{30.87} & \textbf{6.07} & \textbf{0.35} & \textbf{11.79} & \textbf{30.62} & \textbf{6.03} & \textbf{0.34} \\ \midrule

      \multirow{2}{*}{0.5} & \multirow{2}{*}{2} & MonoCon \cite{monocon,choi2023depthdiscriminative}  & \cmark & \cmark &  &  9.44 & 27.37 & 3.71 & 0.14 & 9.37 & 27.17 & 3.68 & 0.14 \\
       & & \ourmethod (Ours) & \cmark & \cmark & \cmark &  \textbf{10.50} & \textbf{30.48} & \textbf{5.52} & \textbf{0.27} & \textbf{10.42} & \textbf{30.24} & \textbf{5.48} & \textbf{0.27} \\
      
      \bottomrule
    \end{tabularx}
    \vspace{0.5em}

    \label{tab:waymo_full}
\end{table*}

\begin{table*}[t]
    \scriptsize
    \caption{\textbf{3D detection accuracy on Waymo (val).} We use the data splits proposed by Reading \etal~\cite{waymo} for training and validation, and report the accuracy for the vehicle class on Level 1 \cite{waymo} (the degree of difficulty). \ourmethod improves over the baseline while being applicable across all camera views in our benchmark: the car view (C), the traffic camera view (T) and the drone view (D). The full evaluation, provided in the \suppmat due to space constraints, is consistent with this conclusion.}
    %$^*$DEVIANT\cite{DEVIANT} uses GUPNet \cite{GUPNet}, which is designed for car view \cite{MONOUNI}. 
    %\setlength \scriptsize
    \centering
    \medskip
    %\resizebox{1.0\textwidth}{!}
    \begin{tabularx}{\linewidth}{
        @{}X c@{\hspace{0.5em}}c@{\hspace{0.5em}}c@{\hspace{0.8em}}
        *{2}{S[table-format=2.2]@{\hspace{1em}}}
        S[table-format=2.2]@{\hspace{0.8em}}
        S[table-format=1.2]@{\hspace{1em}}
        *{2}{S[table-format=2.2]@{\hspace{1em}}}
        S[table-format=1.2]@{\hspace{0.8em}}
        S[table-format=1.2]}
        
    %\begin{tabular}{l | c | c | c | c c c c | c c c c}
      %\Xhline{1pt}
      \toprule
      \multirow{2}{*}{Method} & \multirow{2}{*}{C} & \multirow{2}{*}{T} & \multirow{2}{*}{D} & \multicolumn{4}{c}{$AP_{3D} \uparrow$ $(\text{IoU} = 0.5)$} & \multicolumn{4}{c}{$AP_{3D} \uparrow$ ($\text{IoU} = 0.7$)} \\
      \cmidrule(lr){5-8} \cmidrule(lr){9-12}
      & & & & {All} & {0-30} & {30-50} & {50-$\infty$} & {All} & {0-30} & {30-50} & {50-$\infty$} \\ 
      %\hline
      %\Xhline{1}
      \midrule
      \textbf{With extra data} \\
      Stereoscopic \cite{stereoscopic} & \cmark & & & 7.18 & 22.85 & 5.40 & 1.52 & 1.72 & 6.98 & 1.52 & 0.48 \\
      PatchNet~\cite{PatchNet,PCT} & \cmark & \cmark &  & 2.92 & 10.03 & 1.09 & 0.23 & 0.39 & 1.67 & 0.13 & 0.03  \\
      PCT~\cite{PCT} & \cmark & \cmark & & 4.20 & 14.70 & 1.78 & 0.39 & 0.89 & 3.18 & 0.27 & 0.07 \\
      CaDDN \cite{CaDDN} & \cmark & \cmark & &  17.54 & 45.00 & 9.24 & 0.64 & 4.99 & 14.43 & 1.45 & 0.10 \\
      DID-M3D \cite{DID-M3D}  & \cmark & \cmark & & 20.66 & 40.92 & 15.63 & 5.35 & {---} & {---} & {---} & {---}\\
      TempM3D \cite{tempm3d} & \cmark & \cmark & &  23.95 & 45.46 & 19.50 & 6.31 & {---} & {---} & {---} & {---}\\
      MonoNeRD \cite{mononerd} & \cmark & \cmark & & 31.18 & 61.11 & 26.08 & 6.60 & 10.66 & 27.84 & 5.40 & 0.72 \\
      OccupancyM3D \cite{occupancyM3D} & \cmark & \cmark & & 28.99 & 61.24 & 23.25 & 3.65 & 10.61 & 29.18 & 4.49 & 0.41 \\

      \textbf{Without extra data} \\ 
      M3D-RPN~\cite{M3D-RPN,CaDDN} & \cmark & & & 3.79 & 11.14 & 2.16 & 0.26 & 0.35 & 1.12 & 0.18 & 0.02 \\      

      GUPNet~\cite{GUPNet,DEVIANT} & \cmark & & & 10.02 & 24.78 & 4.84 & 0.22 & 2.28 & 6.15 & 0.81 & 0.03 \\  
      DEVIANT$^*$ ~\cite{DEVIANT} & \cmark & && 10.98 & 26.85 & 5.13 & 0.18  & 2.69 & 6.95 & 0.99 & 0.02  \\
      MonoRCNN++ \cite{MonoRCNN++} & \cmark & & & 11.37 & 27.95 & 4.07 & 0.42 & 4.28 & 9.84 & 0.91 & 0.09 \\
      MonoJSG \cite{MonoJSG} & \cmark & \cmark & & 5.65 & 20.86 & 3.91 & 0.97 & 0.97 & 4.65 & 0.55 & 0.10 \\
      MonoLSS \cite{monolss} & \cmark & \cmark & & 13.49 & 33.64 & 6.45 & 1.29 & 3.71 & 9.82 & 1.14 & 0.16 \\
      MonoUNI \cite{MONOUNI} & \cmark & \cmark & & 10.98 & 26.63 & 4.04 & 0.57 & 3.20 & 8.61 & 0.87 & 0.13 \\
      SSD-MonoDETR \cite{ssd-monodetr} & \cmark & \cmark & & 11.83 & 27.69 & 5.33 & 0.85 & 4.54 & 9.93& 1.18 & 0.15 \\
      MonoXiver-GUPNet \cite{monoxiver} & \cmark & \cmark & & 11.47 & {---} & {---} & {---} & {---} & {---} & {---} & {---} \\      
      MonoXiver-DEVIANT \cite{monoxiver} & \cmark & \cmark & & 11.88 & {---} & {---} & {---} & {---} & {---} & {---} & {---} \\
      MonoCon \cite{monocon} + DDML \cite{choi2023depthdiscriminative} & \cmark & \cmark & & 10.14 & 28.51 & 3.99 & 0.17 & 2.50 & 7.62 & 0.72 & 0.02 \\ \midrule
      MonoCon {\scriptsize (Baseline)} \cite{monocon,choi2023depthdiscriminative} & \cmark & \cmark & & 10.07 & 27.47 & 3.84 & 0.17 & 2.30 & 6.66 & 0.67 & 0.02 \\

      \ourmethod~{\scriptsize (Ours)} & \cmark & \cmark & \cmark & 11.89 & 30.87 & 6.07 & 0.35  & 3.10 & 8.94 & 1.17 & 0.06\\
      \bottomrule

      %\Xhline{1pt}

    \end{tabularx}
    \vspace{0.5em}

    \label{tab:waymo_lvl1}
\end{table*}

\clearpage

\bibliographystyle{splncs04}
\bibliography{arxiv}

\end{document}